\documentclass{article}

\usepackage{preprint}

\usepackage[T1]{fontenc}
\usepackage[utf8]{inputenc}
\usepackage{authblk}
\usepackage{hyperref}       
\usepackage{url}            
\usepackage{booktabs}       
\usepackage{amsfonts}       
\usepackage{nicefrac}       
\usepackage{microtype}      
\usepackage{lipsum}
\usepackage{fancyhdr}       
\usepackage{graphicx}       

\usepackage{graphicx}%
\usepackage{subfigure}
\usepackage{multirow}%
\usepackage{amsmath,amssymb,amsfonts}%
\usepackage{amsthm}%
\usepackage{mathrsfs}%
\usepackage[title]{appendix}%
\usepackage{xcolor}%
\usepackage{textcomp}%
\usepackage{manyfoot}%
\usepackage{booktabs}%
\usepackage{algorithm}%
\usepackage{algorithmicx}%
\usepackage{algpseudocode}%
\usepackage{listings}%
\usepackage{todonotes}

\usepackage{makecell}

\usepackage{colortbl}
\usepackage{xcolor}
\definecolor{tabgray}{gray}{0.90}
\usepackage{enumitem}
\usepackage{comment}
\usepackage{soul}
\usepackage{mathtools}

\usepackage{ulem}
\usepackage{longtable}
\usepackage{multirow}
\usepackage{array}
\usepackage{rotating}
\newcommand{\ours}{DDR}

\usepackage{bm}

\title{Beyond Perceptual Distance: Discrepancy Assessment on Deep Representation for Out-of-Distribution Detection with Diffusion Model}
\author[1,2]{Kun Fang}
\author[3]{Zuopeng Yang}
\author[2]{Haibo Hu}
\author[1]{Xiaolin Huang}
\author[1]{Jie Yang}
\author[4]{Qinghua Tao}

\affil[1]{Department of Automation, Shanghai Jiao Tong University}

\affil[2]{Department of Electrical and
Electronic Engineering, The Hong Kong Polytechnic University
}

\affil[3]{The Intsig Information Co., Ltd., Shanghai, China}

\affil[4]{School of Automation, Beijing Institute of
Technology
}

\begin{document}

\maketitle

\begin{abstract}
Out-of-Distribution (OoD) detection aims to justify whether a given sample is from the training distribution of the classifier-under-protection, i.e., In-Distribution (InD), or from an unknown out distribution.
Recent researches have leveraged Diffusion Models (DMs) for OoD detection due to their powerful distribution modeling capability.
Given an input image, an InD-pretrained DM produces a corresponding InD-aligned counterpart, which serves as a generative reference for comparison.
However, existing DM-based methods typically assess this underlying discrepancy through visual-level distances in the raw image space, which may be misaligned with the distributional discrepancy relevant to OoD detection.
In this work, we investigate the fundamentals of discrepancy assessment in DM-based OoD detection, asking how the discrepancy between an input and its DM-generated counterpart should be formulated, and in which representation spaces and with which metrics it should be measured.
To this end, we propose to assess the discrepancy in a classifier-relative manner by exploiting the representation spaces of the classifier-under-protection, whose training on InD data encodes rich task-relevant InD knowledge.
In particular, we quantify two types of discrepancy: feature-level covariate discrepancy in deep feature representations and logit-level concept discrepancy in output logits, enabling effective differentiation between InD and OoD samples.
Moreover, a subspace-based strategy is devised to refine representations of the DM generation to promote discrepancy assessment.
Together, these designs form our novel detection framework, namely {\ours}.
Extensive experiments on the challenging large-scale ImageNet-1K dataset demonstrate the superior detection performance of {\ours} over both DM-based and non-DM-based methods.
\end{abstract}

\section{Introduction}
\label{sec:intro}
Deep{\let\thefootnote\relax\footnotetext{kun.fang@polyu.edu.hk}} Neural Networks (DNNs) have achieved remarkable success in well-specified settings, i.e., the training and test data are assumed to be independently and identically from the same In-Distribution (InD) \cite{lecun2015deep}. 
However, real-world deployment often exposes DNNs to data from unknown distributions, i.e., Out-of-Distribution (OoD), leading to overconfident yet unreliable predictions with critical safety risks \cite{wu2024unsupervised}. 
Equipping models with abilities to identify OoD samples has therefore become essential for trustworthy AI systems, advancing the development of OoD detection \cite{zhang2023global,yang2024generalized}.

Recent advancements in Diffusion Models (DMs), celebrated for their exceptional ability of modeling complex data distributions and generating high-fidelity images \cite{sohl2015deep,ho2020denoising,song2020score,song2020denoising,dhariwal2021diffusion}, have inspired a new family of OoD detection methods \cite{graham2023denoising,liu2023unsupervised,gao2023diffguard,heng2024out,Abdi_2025_ICCV,yang2024diffusion,du2024dream,pmlr-v267-liao25g,yoon2025diffusion}, as reviewed in Section \ref{sec:related-work-ood-diffusion}. 
Among them, the generation-and-comparison paradigm \cite{graham2023denoising,liu2023unsupervised,gao2023diffguard} is built upon a compelling intuition: 
a DM trained exclusively on InD data can transform an input image $\bm x$ into a counterpart $\bm{\hat x}$ that is inherently biased toward InD, thus providing a generative reference.
If $\bm x$ is truly InD, this generative InD-oriented reference $\bm{\hat x}$ should remain distributionally consistent with $\bm x$; if $\bm x$ is OoD, its reference $\bm{\hat x}$ tends to be drawn toward InD, thus yielding a pronounced discrepancy.
By assessing the discrepancy between $\bm{x}$ and $\bm{\hat x}$, one can thereby infer whether $\bm x$ comes from InD or OoD.

\begin{figure*}[t]
    \centering
    \includegraphics[width=0.9\linewidth]{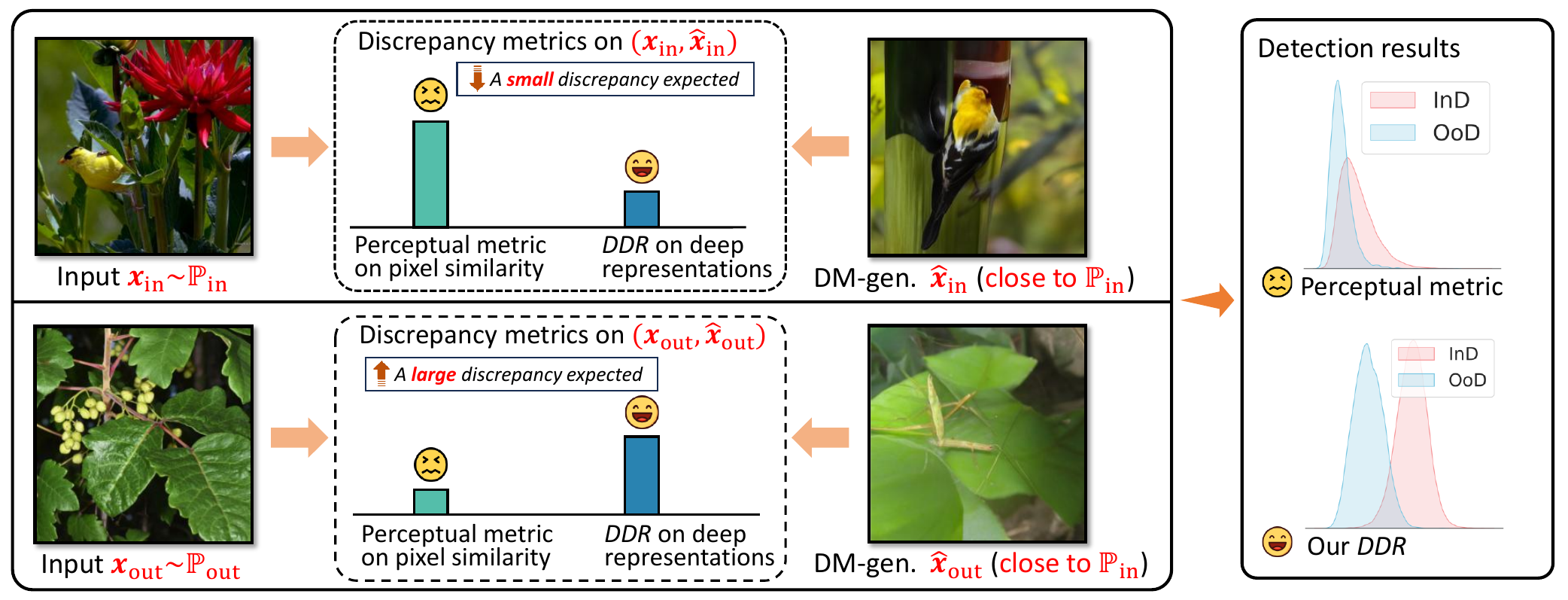}
    \caption{
    Motivation for informative discrepancy assessments. 
    Given an InD image $\bm x_{\rm in}$, its DM generation $\bm{\hat x}_{\rm in}$ is expected to exhibit a small discrepancy with $\bm x_{\rm in}$, whereas the DM-generated counterpart $\bm{\hat x}_{\rm out}$ of an OoD image $\bm x_{\rm out}$ should have a larger discrepancy with $\bm x_{\rm out}$, 
    as the DM is trained to model the distribution w.r.t. InD data.
    However, due to the rich and some unique contents on raw images, a large distance on image pixels can be obtained between $\bm x_{\rm in}$ and $\bm {\hat{x}}_{\rm in}$. 
    Meanwhile, considering the local patterns and textures shared with InD data, it is likely to have a small perceptual distance between $\bm x_{\rm out}$ and $\bm {\hat{x}}_{\rm out}$, leading to unsatisfactory detection results. 
    {\ours} treats the DM-generated image as an InD-oriented reference and assesses whether $\bm x$ and $\bm{\hat x}$ induce consistent representations in the classifier-under-protection.
    }
    \label{fig:intro}
\end{figure*}

However, the way this discrepancy is assessed in existing generation-and-comparison methods is not well aligned with the generative intuition.
The comparison between $\bm{x}$ and its InD-oriented reference $\bm{\hat{x}}$ is typically performed through human-perceptual metrics (e.g., SSIM \cite{wang2004image}, LPIPS \cite{zhang2018unreasonable}, DISTS \cite{ding2020image}) that measure their visual similarity in the raw image space, rather than capturing the distributional discrepancy between $\bm{x}$ and $\bm{\hat{x}}$.
This target mismatch, where the DM provides an InD-oriented reference but the detector evaluates only perceptual similarity, can cause undesirable behavior.
As illustrated in Figure~\ref{fig:intro}, an InD sample with rich and unique image contents may exert a large perceptual distance from its DM-generated counterpart, while an OoD sample with locally similar structures can misleadingly yield little discrepancy.
As a result, perceptual metrics in the raw pixel space may fail to assess the discrepancy truly relevant to OoD detection, limiting the distribution-modeling potential of DMs.
This naturally raises a core question that challenges the foundation of DM-based generation-and-comparison OoD detection:
\begin{center}
{\it How should the discrepancy between $\bm{x}$ and $\bm{\hat{x}}$ be appropriately formulated,\\ and in which representation spaces and with which metrics?}
\end{center}

The key to resolving the above mismatch is to ground discrepancy assessment in the InD knowledge that defines the detection problem.
Such knowledge is not merely conceptual: it is directly tied to the classifier-under-protection, the well-trained DNN that anchors OoD detection and whose training distribution defines InD.
Our answer is therefore to assess the discrepancy in a classifier-relative manner: 
it is no longer a perceptual distance between two images, but an inconsistency between the classifier responses to $\bm x$ and to its InD-oriented reference $\bm{\hat x}$. 
This classifier-relative view naturally directs discrepancy assessment to the representation spaces of the classifier. 
Trained on InD data, its intermediate and output representations encode complementary and informative InD knowledge: 
deep-layer features capture abstract and low-dimensional patterns learned from InD, while output-layer logits reflect the semantic certainty over known InD categories. 
Accordingly, we compare $\bm{x}$ and $\bm{\hat x}$ in these representation spaces, and formulate a novel detection framework that goes beyond direct measurement in raw image pixels.

Concretely, we decompose this classifier-relative relational discrepancy between $\bm x$ and $\bm{\hat x}$ into two complementary aspects, following the established distinction between covariate and concept shifts \cite{tian2021exploring}. 
{\it{(i)}}
The first, feature-level {\it covariate shift}, reflects whether the feature representations of $\bm x$ and $\bm{\hat x}$ deviate under the InD feature covariance: an InD sample and its counterpart should remain close, whereas an OoD sample exhibits a larger deviation.
{\it{(ii)}}
The second, logit-level {\it concept shift}, captures whether $\bm x$ and $\bm{\hat x}$ induce comparable semantic certainty over known classes: an InD sample and its counterpart should align, while an OoD sample typically shows lower certainty than its InD-oriented counterpart.
Accordingly, these two discrepancies are characterized with established yet distinctively applied tools. 
In the feature space, we use the Mahalanobis geometry \cite{mahalanobis2018generalized} induced by InD training features to measure the relational deviation between $\bm{x}$ and $\bm{\hat x}$. 
In the logit space, we use an Energy ratio \cite{lecun2006tutorial} between $\bm{x}$ and $\bm{\hat x}$ to measure relative semantic certainty. 
Importantly, neither metric is used as a standalone detector on $\bm{x}$; both serve for a relational comparison between $\bm{x}$ and its DM-generated counterpart $\bm{\hat x}$.
Moreover, since $\bm{\hat x}$ serves as the InD-oriented reference, we further refine its representations through an InD principal subspace learned from training representations. 
This projection reduces off-subspace variations and generation artifacts that are weakly supported by InD statistics, yielding a reference more faithful to the classifier's InD representation structure. 
We name our proposed  framework as {\bf\ours}, where the {\bf D}iscrepancy induced by {\bf D}iffusion models is captured in deep {\bf R}epresentation spaces.

{\ours} shifts the discrepancy assessment in DM-based OoD detection from image-level perceptual similarity to classifier-relative representation consistency, providing a unified way to exploit both the generative prior of DMs and the discriminative InD knowledge of the deployed classifier.
The superior performance of {\ours} is validated on the challenging large-scale ImageNet-1K dataset \cite{deng2009imagenet} of size $224\times224$. 
The contributions of this work are summarized below.
\begin{itemize}[leftmargin=*]

    \item We identify a target mismatch in existing DM-based generation-and-comparison detectors: they use perceptual distances in raw image space to assess the underlying discrepancy, whereas the desired discrepancy should reflect distributional relational deviation. 
    
    \item We accordingly advocate classifier-relative discrepancy assessment in deep representation spaces through two complementary measures: feature-level covariate shift and logit-level conceptual shift, formulating a novel detection framework {\ours}.
    Mahalanobis distance and Energy ratio are adopted as practical realizations of these two relational discrepancies, respectively.

    \item We introduce a reference refinement strategy that projects representations of the DM-generated counterpart onto an InD principal subspace, improving the faithfulness of the generated reference to the classifier's InD representation structure and thereby enhancing discrepancy assessment.
\end{itemize}

In the remainder, Sections \ref{sec:preliminary} and \ref{sec:related-work-ood-diffusion} outline preliminaries and related work for OoD detection and diffusion models.
Section \ref{sec:method} elaborates our {\ours} framework with technical details and explanations.
Empirical results are presented in Section \ref{sec:exp}.
Conclusions and discussions are enclosed in Section \ref{sec:conclusion}.

\section{Preliminaries}
\label{sec:preliminary}

\subsection{Out-of-distribution Detection}
Given an input $\bm {x}\in\mathbb{R}^d$, OoD detection \cite{yang2024generalized} is formulated as a bi-classification task with a scoring function $S(\cdot)$:
\begin{equation}
\label{eq:ood-bi-classification}
{\rm category\ of}\ \bm {x}=
\left\{
\begin{array}{ll}
     \mathrm{InD},& S(\bm {x})>s,\\
     \mathrm{OoD},&S(\bm {x})<s.
\end{array}
\right.
\end{equation}
If the detection score $S(\bm {x})$ is greater than the threshold $s$, $\bm {x}$ is classified as InD, and vice versa.
The threshold $s$ is determined such that most of the InD samples are correctly classified.
The core challenge therefore lies in designing a scoring function $S(\cdot)$ that reliably captures the discrepancy between InD and OoD data. 

\subsection{Diffusion Models}

Diffusion Models (DMs), or saying diffusion denoising probabilistic models, have achieved great successes in generating high-fidelity images from the learned distribution \cite{sohl2015deep,ho2020denoising,song2020score,song2020denoising,dhariwal2021diffusion}.
In DMs, the so-called {\it forward} and {\it reverse} diffusion processes are involved, where a generated image can be obtained for any given input image.
\begin{itemize}[leftmargin=*]
    \item {Forward} process: an image $\bm {x}$ is corrupted with $T$ steps of Gaussian noises until it turns into pure noises $\bm {x}_T$:
    \begin{equation}
    \label{eq:dm-forward}
    q(\bm {x}_t|\bm {x}_{t-1})=\mathcal{N}(\bm {x}_t|\sqrt{1-\beta_t}\bm {x}_t,\beta_t{\bf {I}}),
    \end{equation}
    where $0\leq t\leq T$ and $\beta_t$ follows a fixed variance schedule so that $\bm {x}_T$ is close to an isotropic Gaussian $\mathcal{N}(\bm {0},{\bf {I}})$.
    \item {Reverse} process: a DNN with parameters $\bm {\theta}$ is introduced to remove the noises in $\bm {x}_T$ via multiple steps, until a clean image $\bm {\hat x}=p_{\bm {\theta}}(\bm {x}_0|\bm {x}_T)$ is finally recovered:
    \begin{equation}
    \label{eq:dm-reverse}
    p_{\bm {\theta}}(\bm {x}_{t-1}|\bm {x}_t)=\mathcal{N}(\bm {x}_{t-1}|\bm {\mu}_{\bm {\theta}}(\bm {x}_t,t),\bm {\Sigma}_{\bm {\theta}}(\bm {x}_t,t)),
    \end{equation}
    where $\bm {\mu}_{\bm {\theta}}$ and $\bm {\Sigma}_{\bm {\theta}}$ are mean and variances of Gaussian noises learned from the DNN model with network parameters $\bm {\theta}$.
\end{itemize}

\section{Related work}
\label{sec:related-work-ood-diffusion}

The remarkable ability of DMs to model data distributions and generate realistic images has recently inspired a variety of novel OoD detection methods \cite{graham2023denoising,liu2023unsupervised,gao2023diffguard,heng2024out,Abdi_2025_ICCV,yang2024diffusion,du2024dream,pmlr-v267-liao25g,yoon2025diffusion}. 
We outline three mainstream paradigms in this direction, each driven by a distinct rationale, and refer to Appendix \ref{app:sec:related-work} for other non-DM-based detection approaches.

\begin{itemize}[leftmargin=*]
    \item {\bf Generation-and-Comparison.} Given an input image $\bm x$ from an unknown distribution, a series of methods \cite{graham2023denoising,liu2023unsupervised,gao2023diffguard} perform the complete forward and reverse processes on $\bm x$ (cf. Eqns.\eqref{eq:dm-forward} and \eqref{eq:dm-reverse}) to obtain its DM generation $\bm{\hat x}$.
    Detection is then based on quantifying the underlying discrepancy between $\bm x$ and $\bm{\hat{x}}$. 
    The core rationale is that for an InD input ${\bm x}_{\rm in}$, the DM (trained on InD data) should generate its counterpart ${\bm{\hat{x}}}_{\rm in}$ that lies near InD and thus also remains in close proximity to ${\bm x}_{\rm in}$. 
    For an OoD input ${\bm x}_{\rm out}$, however, the DM's generative prior dominates, pulling $\bm{\hat{x}}_{\rm out}$ toward the InD and thus creating discrepancy with the original ${\bm x}_{\rm out}$. 
    
    The pioneering work \cite{graham2023denoising} corrupts the given image $\bm {x}$ at a range of different noise levels and generates the recovered images $\bm {\hat x}$. 
    Then in LMD \cite{liu2023unsupervised}, it masks $\bm {x}$ in order to lift it off its distribution, and employs DMs to inpaint the masked image.
    Another work DiffGuard \cite{gao2023diffguard} develops multiple test-time techniques in the reverse process of DMs to generate images $\bm {\hat x}$ with larger semantic differences from original images $\bm {x}$.
    Nevertheless, these methods ultimately resort to perceptual metrics, such as SSIM \cite{wang2004image}, LPIPS \cite{zhang2018unreasonable} and DISTS \cite{ding2020image}, to quantify human-perceived similarity in the raw image contents between $\bm x$ and $\bm{\hat x}$, which do not necessarily capture their underlying discrepancy, as shown with the counterexamples in Figure \ref{fig:intro}.
    Our {\ours} falls in this category and focuses on the discrepancy assessment stage.

    \item {\bf Trajectory-Monitoring.} In contrast to output-based comparisons, one can also explore the internal dynamics of the diffusion process itself for OoD detection, as done in \cite{heng2024out,Abdi_2025_ICCV,yang2024diffusion}.
    The underlying hypothesis is that the DM, being specialized for InD data, produces trajectories or estimates for OoD inputs that are statistically inconsistent with those for InD data.
    Methods such as DiffPath \cite{heng2024out} and its subsequent work \cite{Abdi_2025_ICCV} monitor the DM's internal signals, i.e., predictive scores and estimated noise at each diffusion step given the input $\bm x$, and then fit a kernel density estimation model on these signals' statistics for OoD detection.
    Another work \cite{yang2024diffusion} proposes to train an additional network to reconstruct the intermediate features from the DM across time steps for OoD detection.
    This paradigm differs from our work by probing the DM's internal states rather than its outputs.

    \item {\bf Outlier-Synthesis-for-Training.} 
    DMs can also be deployed for data augmentation to facilitate OoD detection \cite{du2024dream,pmlr-v267-liao25g,yoon2025diffusion}.
    In this line of works, it targets on generating diverse and realistic OoD outliers to augment the training set, thereby regularizing the classifier to improve its inherent robustness against OoD through (semi-)supervised model training, instead of unsupervised discrepancy assessment on images or diffusion states. 
    This paradigm requires not only augmented outlier data, but also the optimization on the classifier, similar to adversarial training. 
    Thus, its working mechanism is orthogonal to our work and also the aforementioned two paradigms.
\end{itemize}

\section{Methodology}
\label{sec:method}

Diffusion models bring new potentials for OoD detection, complementing to the generic discriminative detection methods which only leverage the responses from the classifier-under-protection. 
Existing generation-and-comparison DM-based detection methods \cite{graham2023denoising,liu2023unsupervised,gao2023diffguard} focus on techniques of improving the DM generation $\bm{\hat x}$, and resort to use perceptual distances to measure the visual similarity between $\bm x$ and $\bm{\hat x}$, as summarized in Sec.\ref{sec:related-work-ood-diffusion}.
We argue that this assessment is mismatched with the underlying discrepancy truly relevant to OoD detection. 
By going beyond the straightforward assessment on pixel-wise distance between raw images, our {\ours} framework advances to address the fundamental question raised in Sec.\ref{sec:intro}: effective discrepancy assessment between $\bm x$ and $\bm{\hat x}$ should be executed in a classifier-relative manner, particularly in representation spaces well encoded with InD knowledge, i.e., the deep representation spaces learned by the classifier-under-protection $f(\cdot)$ which is well optimized on InD training data.
In the remainder, investigations into deep representation spaces are presented in Sec.\ref{sec:method-analysis} to reveal the distinct discrepancy in representations of $\bm x$ and $\bm{\hat x}$, and corresponding evaluations on such discrepancy are derived in Sec.\ref{sec:method-metrics}. 

\noindent{\bf Notations.}\quad 
Given an image $\bm {x}\in\mathbb{R}^d$ to detect and a DM pretrained on InD data, $\bm {x}$ is corrupted via adding noises (cf. Eqn.\eqref{eq:dm-forward}) and then gets denoised (cf. Eqn.\eqref{eq:dm-reverse}), yielding its DM generation $\bm {\hat x}\in\mathbb{R}^d$. 
In OoD detection, it  tells whether a given image $\bm {x}$ is from the
same distribution (InD) with the training set of a DNN $f(\cdot):\mathbb{R}^d\rightarrow\mathbb{R}^c$, a.k.a., the classifier-under-protection. 
$f(\cdot)$ is readily available in OoD detection, where we denote the feature representations learned in the penultimate layer as $\bm {h}_{\bm {x}},\bm {h}_{\bm {\hat x}}\in\mathbb{R}^m$ w.r.t. $\bm x, \bm {\hat x}$ and their corresponding $c$-class logits  in the output layer as $\bm {z}_{\bm {x}},\bm {z}_{\bm {\hat x}}\in\mathbb{R}^c$.

\subsection{Investigations into deep representation spaces}
\label{sec:method-analysis}

Both the DM and classifier $f(\cdot)$ are optimized to encode the distributional information w.r.t. InD, so that the given InD image ${\bm x}_{\rm in}$ and it DM-generated counterpart $\bm{\hat x}_{\rm in}$ can lie closely to each other in the representation space learned by $f(\cdot)$.
In contrast, for an OoD ${\bm x}_{\rm out}$ and its DM generation ${\bm{\hat x}}_{\rm out}$, their representations from $f(\cdot)$ are deviated from each other, since  ${\bm{\hat x}}_{\rm out}$ is generated to be aligned with InD through the DM.
Such representation-level discrepancy between $\bm x$ and $\bm{\hat x}$ appears differently across the network layers of $f(\cdot)$. 
In particular, we investigate the covariate shift and concept shift in the feature and logit spaces, as pointed out in \cite{tian2021exploring} for OoD data.
In the deep feature space, a covariate shift manifests as a geometric misalignment among the representations of InD and OoD samples, as the classifier has never seen OoD data in its optimization. 
In the logit space, a concept shift is evident, as reflected in their semantic uncertainty: InD logits are typically dominated by one known class, whereas OoD logits tend toward a uniform distribution due to categorical ambiguity.
We provide empirical evidences that the representations of $\bm x$ and $\bm{\hat x}$ exhibit both covariate and concept shifts, motivating our design of evaluation metrics tailored correspondingly in addressed representation spaces.

\begin{figure*}[t]
    \centering
    \includegraphics[width=0.95\linewidth]{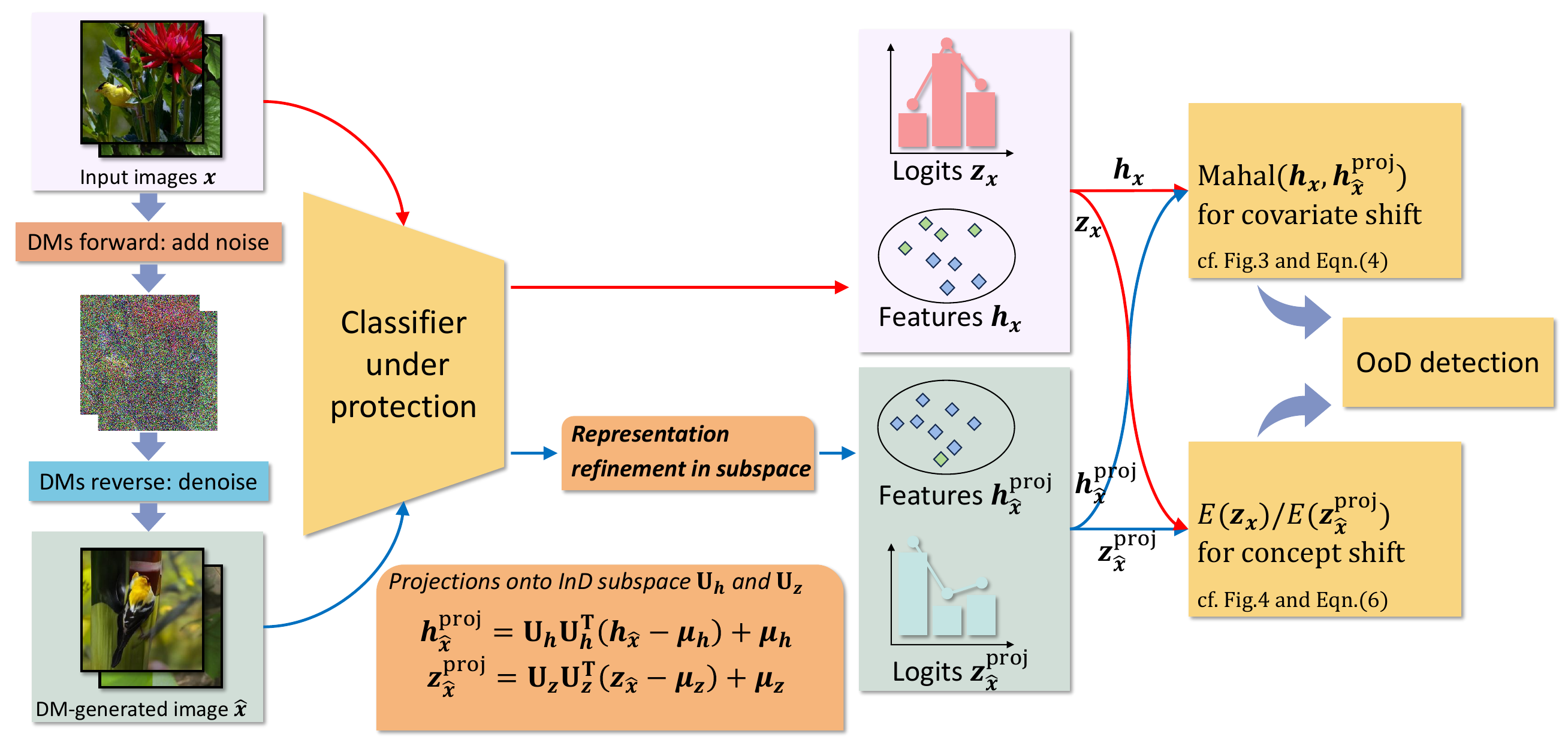}
    \caption{Schematic illustration of \ours. The classifier-under-protection is introduced to assess the representation discrepancy between an input image $\bm {x}$ and its DM generation $\bm {\hat x}$ in its feature space and logit space. A representation refinement strategy is applied to the representations of $\bm {\hat x}$ via subspace projection to facilitate discrepancy assessment and boost OoD detection.}
    \label{fig:framework}
\end{figure*}

\begin{figure*}[t]
    \centering
    \includegraphics[width=0.99\linewidth]{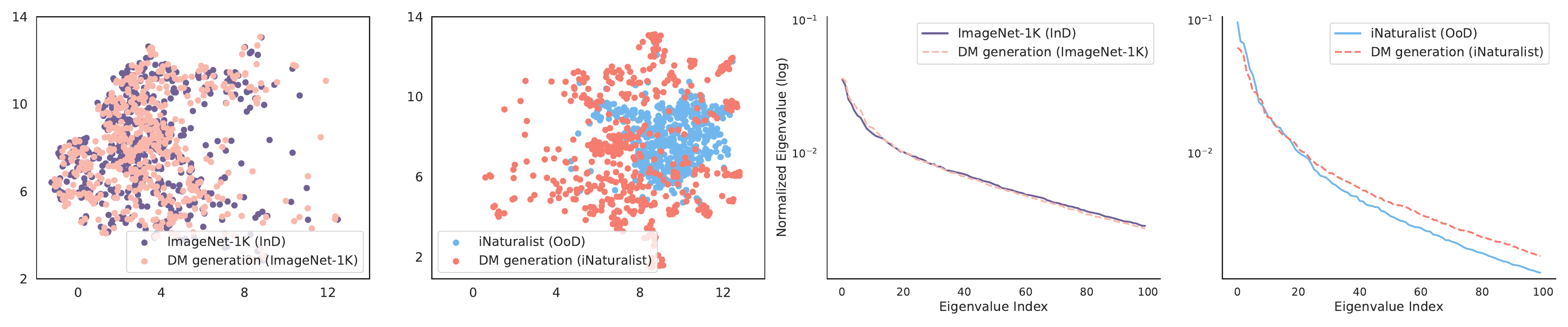}
    \caption{UMAP visualization \cite{mcinnes2018umap} and normalized eigenvalue spectra of input features ${\bm h}_{\bm x}$ and their DM-generated counterparts ${\bm h}_{\bm{\hat{x}}}$.
    {\bf From left to right}: UMAP embeddings of InD features (ImageNet-1K);
    UMAP embeddings of OoD features (iNaturalist);
    top-100 normalized eigenvalues of InD features (ImageNet-1K);
    top-100 normalized eigenvalues of OoD features (iNaturalist).
    For clarity, the two groups of UMAP and eigenvalues are displayed in separate yet aligned panels, with both sharing the same axis scale.}
    \label{fig:analys-feat}
\end{figure*}

\paragraph{Covariate shift in feature space}
In DNNs, covariate shifts manifest misalignments in the geometric structure of the learned features.
For illustrations with OoD setups, we investigate the network of $f(\cdot)$ with two groups of features: {\it (i)} the features ${\bm h}_{{\bm x}_{\rm in}}$ of InD images ${\bm x}_{\rm in}$ and the features ${\bm h}_{{\bm{\hat x}}_{\rm in}}$ of the corresponding DM-generated counterparts ${\bm{\hat x}}_{\rm in}$, {\it (ii)} the features ${\bm h}_{{\bm x}_{\rm out}}$ of OoD images ${\bm x}_{\rm out}$ and the features ${\bm h}_{{\bm{\hat x}}_{\rm out}}$ of their DM generation ${\bm{\hat x}}_{\rm out}$. 
The two groups of features are jointly processed via UMAP \cite{mcinnes2018umap} and are visualized in two separate yet aligned panels in Figure \ref{fig:analys-feat}; we additionally plot the top-$K$ normalized eigenvalues $\tilde{\lambda}_k = \lambda_k / \sum_j \lambda_j$ of the corresponding feature covariance matrices to quantify the principal geometric structure of each distribution.
These visualizations reveal a discernible phenomenon on their relational discrepancy:
\begin{itemize}[leftmargin=*]
    \item For InD images, the features ${\bm h}_{{\bm x}_{\rm in}}$ and ${\bm h}_{{\bm{\hat x}}_{\rm in}}$ lie in close proximity as expected, since  ${\bm x}_{\rm in}$ and its DM generation ${\bm{\hat x}}_{\rm in}$ can be well encoded by $f(\cdot)$  that is learned with InD training data (1st panel in Figure \ref{fig:analys-feat}). 
    Consistently, the eigenvalue spectra of ${\bm h}_{{\bm x}_{\rm in}}$ and ${\bm h}_{{\bm{\hat x}}_{\rm in}}$ exhibit closely aligned decay profiles, confirming that DM generation preserves the principal geometric structure of the InD feature distribution (3rd panel in Figure~\ref{fig:analys-feat}).
    \item For OoD images, there exists an evident geometric shift between ${\bm h}_{{\bm x}_{\rm out}}$ and ${\bm h}_{{\bm{\hat x}}_{\rm out}}$, such that the features ${\bm h}_{{\bm x}_{\rm out}}$ lie far from the InD feature cluster but the features ${\bm h}_{{\bm{\hat x}}_{\rm out}}$ of DM generation appear much closer, since the DM is trained to generate from InD (2nd panel in Figure \ref{fig:analys-feat}). 
    This shift is further corroborated by the eigenvalue spectra: the spectrum of ${\bm h}_{{\bm{\hat x}}_{\rm out}}$ deviates noticeably from that of ${\bm h}_{{\bm x}_{\rm out}}$ and migrates toward the InD eigenvalue profile, reflecting the tendency of the DM to pull OoD generations onto the InD manifold (4th panel in Figure~\ref{fig:analys-feat}). 
    This spectral misalignment constitutes a quantitative signature of covariate shift and provides the discriminative basis for OoD detection.
\end{itemize}

\begin{figure*}[t]
    \centering
    \includegraphics[width=0.99\linewidth]{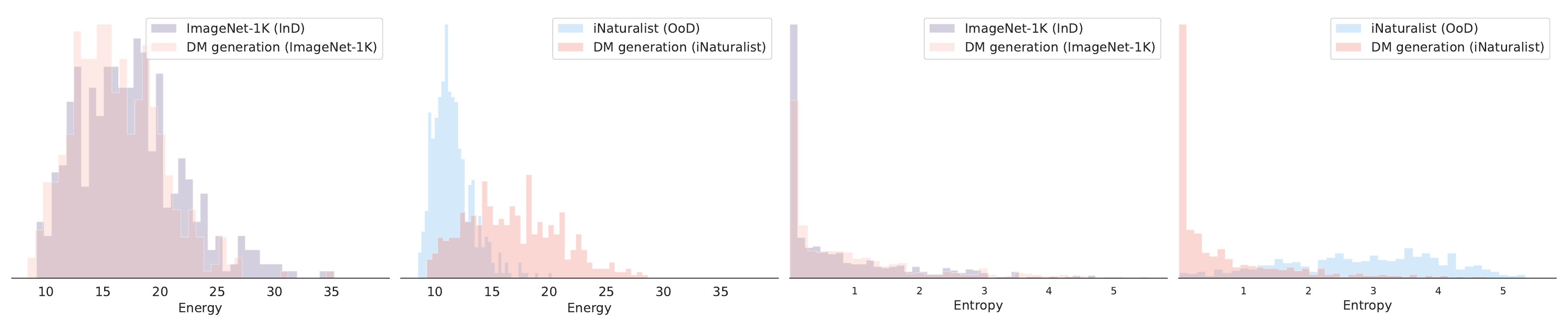}
    \caption{Histograms of Energy and Entropy values of input logits ${\bm z}_{\bm x}$ and the corresponding DM generation logits ${\bm z}_{\bm{\hat x}}$.
    {\bf From left to right}: Energy histogram of InD (ImageNet-1K); Energy histogram of OoD (iNaturalist); Entropy histogram of InD (ImageNet-1K); Entropy histogram of OoD (iNaturalist).
    For clarity, the two logit groups are displayed in separate yet aligned panels for each metric, with both sharing the same axis scale within each metric.}
    \label{fig:analys-logit}
\end{figure*}

\paragraph{Concept shift in logit space}
In DNNs, InD and OoD data also pertain deviations in semantic contents, termed as concept shifts, which are unveiled in logit space with classification probabilities in the output layer.
For an InD image ${\bm x}_{\rm in}$, the classifier typically produces the logit vector ${\bm z}_{{\bm x}_{\rm in}}$ with one dominant dimension, indicating the high confidence for the classification category. 
For OoD samples ${\bm x}_{\rm out}$, the semantic categories are absent during training, incurring uniform-like logits ${\bm z}_{{\bm x}_{\rm out}}$, reflecting the classification uncertainty.
To evaluate such concept shifts, we consider two statistics derived from logits $\bm z_{\bm x}$.
The first is the Energy function \cite{lecun2006tutorial}, which serves as a density-aware measure of semantic certainty:
\begin{equation}
\label{eq:energy}
E({\bm z_{\bm x}})=\log\sum\nolimits_{i=1}^ce^{{\bm z}_{{\bm x},i}}.
\end{equation}
The softmax probability of $\bm x$ for class $i$ is ${\bm p}_{{\bm x},i}=e^{{\bm z}_{{\bm x},i}}/{\sum_i^ce^{{\bm z}_{{\bm x},i}}}$, so that $E({\bm z_{\bm x}})$ reflects both the magnitude and concentration of logits: a sharply concentrated value on one class (high certainty) yields a high $E({\bm z_{\bm x}})$, whereas a uniform-like distribution (high uncertainty) leads to a low $E({\bm z_{\bm x}})$.
The second is the Shannon Entropy of the predicted distribution, which directly quantifies the spread of classification probabilities independently of their magnitude:
\begin{equation}
\label{eq:entropy}
H({\bm p_{\bm x}})=-\sum\nolimits_{i=1}^c {\bm p}_{{\bm x},i}\log {\bm p}_{{\bm x},i}.
\end{equation}
While $E(\cdot)$ is sensitive to the overall scale of logits, $H(\cdot)$ is scale-invariant and measures the distributional uncertainty after softmax normalization; together they provide complementary evidence of concept shift.
In Figure \ref{fig:analys-logit} we present histograms of both Energy and Entropy for two groups of logits w.r.t.\ InD and OoD images and their DM generations.
\begin{itemize}[leftmargin=*]
    \item For InD images, the Energy distributions of ${\bm z}_{{\bm x}_{\rm in}}$ and ${\bm{\hat z}}_{{\bm x}_{\rm in}}$ overlap and exhibit high values, since the InD-trained classifier has high semantic certainty for both ${\bm x}_{\rm in}$ and ${\bm{\hat x}}_{\rm in}$ (1st panel in Figure \ref{fig:analys-logit}). Consistently, the Entropy distributions of the two groups remain low and closely aligned, confirming that the probability mass stays concentrated on a single predicted class for both the original and DM-generated InD images (3rd panel in Figure \ref{fig:analys-logit}).

    \item For OoD inputs, the logits ${\bm z}_{{\bm x}_{\rm out}}$ show notably lower Energy values, indicating high uncertainty with the InD-trained classifier, whereas the DM generation ${\bm{\hat{x}}}_{\rm out}$ produces logits ${\bm z}_{{\bm{\hat x}}_{\rm out}}$ with higher certainty, since the DM is optimized to align with the InD prior for generation (2nd panel in Figure \ref{fig:analys-logit}). The Entropy histograms further corroborate this shift from a distributional perspective: ${\bm z}_{{\bm x}_{\rm out}}$ exhibits markedly higher Entropy, reflecting the spread-out, uniform-like classification probabilities for OoD content, while the Entropy of ${\bm z}_{{\bm{\hat x}}_{\rm out}}$ drops toward the InD range, as the DM-regenerated images carry InD-like semantics that the classifier can more confidently categorize (4th panel in Figure \ref{fig:analys-logit}).
\end{itemize}

The evidences in Figures \ref{fig:analys-feat} and  \ref{fig:analys-logit} demonstrate a striking parallel to  the covariate shift and concept shift between InD and OoD data with their DM generation, as manifested in deep feature space and logit space of the classifier-under-protection.
The informativeness of incorporating the covariate and concept shifts motivates  to leverage deep representations for DM-based OoD detection and devise effective metrics in such representation spaces, leading to our proposed new framework {\ours} for OoD detection.

\subsection{Evaluations in deep representation spaces}
\label{sec:method-metrics}

In the previous section, {\ours} reveals a pronounced relational discrepancy between $\bm x$ and $\bm{\hat x}$ in the classifier's deep feature and logit spaces, reflecting covariate and concept shifts, respectively. 
Here, we demonstrate that such discrepancy can be sufficiently and effectively captured by well-established metrics, simply by redirecting them from standalone scoring to relational comparison within the {\ours} framework. 
This design reflects a key insight of our work: when suitably adapted to relational comparison, even classical tools can richly characterize the distributional discrepancy relevant to OoD detection. 
We also note that the {\ours} framework is compatible with more advanced metrics that may further improve performance, an avenue we leave open for future exploration.

\paragraph{{\ours} for covariate shift in feature space} 
For deep features ${\bm h}_{\bm x}$ and ${\bm h}_{\bm{\hat x}}$, we consider the Mahalanobis distance \cite{mahalanobis2018generalized} as the core metric for this  representation space:
\begin{equation}
\label{eq:ddr-feat}
\epsilon_{\rm feat}(\bm {x},\bm {\hat x})
=\sqrt{(\bm{h}_{\bm {x}}-\bm{h}_{\bm{\hat x}})^\top{\rm\bf\Sigma}^{\dagger}_{\bm{h}}(\bm{h}_{\bm {x}}-\bm{h}_{\bm{\hat x}})},
\end{equation}
where ${\rm\bf\Sigma}_{\bm{h}}$ denotes the covariance matrix of the InD training features also from the classifier-under-protection and $\dagger$ denotes its pseudo-inverse.

The Mahalanobis distance provides a geometric measure that accounts for the inherent correlation structure between ${\bm h}_{\bm x}$ and ${\bm h}_{\bm{\hat x}}$ under the training feature covariance ${\rm\bf\Sigma}_{\bm{h}}$.
To be specific, the Mahalanobis distance gives less weight to directions of high variance and more weight to directions of low variance when measuring the distance between $\bm h_{\bm x}$ and ${\bm h}_{\bm{\hat x}}$.
This makes it a far more sensitive metric for geometric misalignment than the naive Euclidean distance ($\|\bm{h}_{\bm {x}}-\bm{h}_{\bm{\hat{x}}}\|_2$), as the latter implicitly assumes all feature dimensions are isotropic and equally important, which is hardly satisfied in real-world data.

Within our detection framework {\ours}, $\epsilon_{\rm feat}$ evaluates the deviation of $\bm{h}_{\bm {x}}$ from InD in the feature space, since the InD information has been effectively charaterized by the InD-trained classifier $f$ within the training feature covariance ${\rm\bf\Sigma}_{\bm{h}}$ and features $\bm{h}_{\bm{\hat x}}$ of the DM-generated $\bm{\hat x}$.
Consequently, a small Mahalanobis distance implies that $\bm{h}_{\bm {x}}$ is geometrically consistent with InD in the feature space, suggesting it is more likely to come from InD, while a large distance implies a significant covariate shift, indicating that $\bm x$ possibly originates from an out distribution whose feature geometry is misaligned with the learned InD structure.

\paragraph{{\ours} for concept shift in logit space} 
The Energy function (Eqn.\eqref{eq:energy}) itself serves as a natural detector for concept shift between InD and OoD data as shown in Figure \ref{fig:analys-logit}, and has been leveraged in prior non-DM-based OoD detection methods \cite{liu2020energy,park2023nearest}, where it is directly applied to the logits ${\bm z}_{\bm x}$.
Our {\ours} appears a fundamentally different framework, as we are not evaluating the absolute semantic uncertainty of $\bm x$ in isolation.
Instead, we have a DM-generated reference $\bm{\hat x}$ aligned with InD  as a relative measure, since its Energy $E({\bm z}_{\bm{\hat x}})$ under the InD-trained classifier provides a calibrated reference of InD certainty.
An InD sample ${\bm x}_{\rm in}$ produces Energy $E({\bm z}_{\bm{x}_{\rm in}})$ closely matching this baseline, while Energy of an OoD sample ${\bm x}_{\rm out}$ can give  distinctive discrepancy from $E({\bm z}_{\bm{\hat x}_{\rm out}})$.
Thus, we devise the Energy ratio of ${\bm z}_{\bm{x}}$ over ${\bm z}_{\bm{\hat x}}$ as the discrepancy metric of {\ours} in the logit space:
\begin{equation}
\label{eq:ddr-prob}
\epsilon_{\rm logt}(\bm {x},\bm {\hat x})
= {E({\bm z}_{\bm{x}})} / {E({\bm z}_{\bm{\hat x}})},
\end{equation}
which quantifies the relative semantic certainty of the input sample $\bm x$ compared to its InD-oriented reference $\bm {\hat x}$. 
A ratio less than 1 implies lower certainty for $\bm x$, serving as a strong indicator of concept shift to OoD, and vice versa.

\subsection{Representation refinement in subspace}
\label{sec:method-subspace}
With our {\ours} framework, we further enhance the discrepancy assessment by refining representations of the DM generation $\bm{\hat x}$, such that their alignment with InD gets improved.
To this end, we leverage Principal Component Analysis (PCA) \cite{pearson1901liii,abdi2010principal} to obtain a principal subspace which is learned from the InD training data. 
Through the projections onto this InD subspace, the representations of  DM generation $\bm{\hat x}$ can be refined by diminishing the potential anomalous information from OoD, thereby forcing them to adhere more closely to InD. 
In this way, the refined representations of $\bm x$ can even exhibit a sharp contrast to that of $\bm x$, which further promotes their discrepancy assessment.

Given the features $\{{\bm h}_i^{\rm tr}\}_{i=1}^{N_{\rm tr}}$ and logits $\{{\bm z}_i^{\rm tr}\}_{i=1}^{N_{\rm tr}}$  from the InD-trained classifier $f(\cdot)$ on training data $\{{\bm x}_i^{\rm tr}\}_{i=1}^{N_{\rm tr}}$, we construct the principal subspace from their induced covariance matrices: 
\begin{equation}
\begin{aligned}
\label{eq:proj}
{\bf\Sigma}_{\bm h}
=\sum\nolimits_{i=1}^{N_{\rm tr}}\left({\bm h}_i^{\rm tr}-{\bm\mu}_{\bm h}
\right)\left({\bm h}_i^{\rm tr}-{\bm\mu}_{\bm h}\right)^\top,
\quad
{\bf\Sigma}_{\bm z}
=\sum\nolimits_{i=1}^{N_{\rm tr}}\left({\bm z}_i^{\rm tr}-{\bm\mu}_{\bm z}
\right)\left({\bm z}_i^{\rm tr}-{\bm\mu}_{\bm z}\right)^\top,
\end{aligned}
\end{equation}
where ${\bf\Sigma}_{\bm h}\in\mathbb{R}^{m\times m}$ and ${\bf\Sigma}_{\bm z}\in\mathbb{R}^{c\times c}$ are the covariances on features and logits with the means ${\bm\mu}_{\bm h}=\frac{1}{N_{\rm tr}}\sum\nolimits_{i=1}^{N_{\rm tr}}{\bm h}_i^{\rm tr}$ and ${\bm\mu}_{\bm z}=\frac{1}{N_{\rm tr}}\sum\nolimits_{i=1}^{N_{\rm tr}}{\bm z}_i^{\rm tr}$, respectively. 
With PCA techniques, principal components are computed by eigendecomposition to the covariance matrices, such that the top eigenvectors span the principal subspace. 
We choose the subspace that covers over 90\% explained variance and gets spanned  by the corresponding top eigenvectors ${\bf U}_{\bm h}\in\mathbb{R}^{m\times {q_h}}$ and  ${\bf U}_{\bm z}\in\mathbb{R}^{c\times {q_z}}$  w.r.t. ${\bf\Sigma}_{\bm h}\in\mathbb{R}^{m\times m}$ and ${\bf\Sigma}_{\bm z}\in\mathbb{R}^{c\times c}$. In this way, such subspaces capture the most informative pattern from the InD training data and meanwhile assists potential anomalous OoD removal in {\ours}.

This subspace-based strategy for projected representations is deployed at the inference stage, in which we conduct the representation refinement by reconstructions of the features ${\bm h}_{\bm{\hat x}}$ and logits ${\bm z}_{\bm{\hat x}}$ in their corresponding subspaces through projection matrices ${\bf U}_{\bm h}$ and ${\bf U}_{\bm z}$, respectively, such that:
\begin{equation}
\begin{aligned}
{\bm h}_{\bm{\hat x}}^{\rm proj} 
={\bf U}_{\bm h}{\bf U}_{\bm h}^\top({\bm h}_{\bm{\hat x}}-{\bm\mu}_{\bm{h}})+{\bm\mu}_{\bm h},\quad
{\bm z}_{\bm{\hat x}}^{\rm proj}
={\bf U}_{\bm z}{\bf U}_{\bm z}^\top({\bm z}_{\bm{\hat x}}-{\bm\mu}_{\bm{z}})+{\bm\mu}_{\bm z}.
\end{aligned}
\end{equation}
The refined feature and logit representations w.r.t. DM-generated counterparts, i.e., ${\bm h}_{\bm{\hat x}}^{\rm proj}$ and ${\bm z}_{\bm{\hat x}}^{\rm proj}$, are thereby imposed with improved alignment with InD in the principal subspace learned by InD data. 
In this way, the discrepancy between the representations of an input image ${\bm h}_{\bm{x}}$ and ${\bm z}_{\bm{x}}$ and its refined  ${\bm h}_{\bm{\hat x}}^{\rm proj}$ and ${\bm z}_{\bm{\hat x}}^{\rm proj}$ from DM generation gets narrowed with InD inputs, but is pronounced with OoD inputs.
Thus, this strategy for refined representations lead to further enhancement of the discrepancy measurement that well differentiates InD data from OoD.

In {\ours}, we leverage both the deep feature space and the logit space with the classifier-under-protection, and consider both covariate shift and concept shift therein by accessing the discrepancy with the DM-generated counterparts, forming a unified detection score:
\begin{equation}
\begin{aligned}
\label{eq:ddos-final}
S_{\mathrm{DDR}}(\bm {x})
=\frac{\lambda}{\epsilon_{\rm feat}}+(1-\lambda)\cdot{\epsilon_{\rm logt}}
=\frac{\lambda}{\sqrt{(\bm{h}_{\bm {x}}-\bm{h}^{\rm proj}_{\bm{\hat x}})^\top{\rm\bf\Sigma}^{\dagger}_{\bm{h}}(\bm{h}_{\bm {x}}-\bm{h}^{\rm proj}_{\bm{\hat x}})}}
+(1-\lambda)\frac{E({\bm z})}{E({\bm z}^{\rm proj}_{\bm{\hat x}})}.
\end{aligned}
\end{equation}
$\lambda \in [0, 1]$ is a coefficient that balances the effect of each metric. 
The reciprocal term on $\epsilon_{\rm feat}$ ensures that the final score $S_{\rm\ours}(\bm{x})$ becomes lower for OoD samples and higher for InD samples, so as to align with standard detection scoring.

\noindent{\bf Remark.}\quad
We highlight the significant distinctiveness of our {\ours} compared to prior OoD detection methods that also utilize the Mahalanobis distance \cite{lee2018simple,mullermahalanobis++} and the Energy function \cite{liu2020energy} but in a rather different manner.
The fundamental difference lies in their detection paradigms: through a InD-pretrained DM, {\ours} generates a reference $\bm{\hat x}$ from the input $\bm x$ and assesses the discrepancy in representation spaces between $\bm x$ and $\bm{\hat x}$ with such metrics for OoD detection.
In contrast, other related methods do not consider generative counterparts as references and solely rely on the classifier-under-protection by directly measuring its network responses with such metrics.

\section{Experiments}
\label{sec:exp}
In this section, extensive experiments are conducted to evaluate the proposed {\ours} with comparisons to diverse existing detection methods. 
The source code for {\ours} has been publicly released\footnote{\href{https://github.com/fanghenshaometeor/ood-ddr}{https://github.com/fanghenshaometeor/ood-ddr}}.
We also report implementation details of {\ours} in Appendix \ref{app:sec:implementation}.

\subsection{Setups}
\label{sec:exp-setups}

\noindent\textbf{Benchmarks.}
In experiments, we primarily use the large-scale ImageNet-1K dataset \cite{deng2009imagenet} with an image size of $224\times224$ as the InD.
We highlight the importance of this InD dataset setup for two reasons.
First, the powerful generative capacity of DMs \cite{dhariwal2021diffusion} is most fully realized on large size images like those in ImageNet-1K; applying DMs to small-scale images such as $32\times32$ CIFAR is somehow outdated and ill-suited in the modern era.
Second, the complexity and scale of ImageNet-1K provide a challenging testbed for evaluating DM-based OoD detection methods, one that remains underexplored in most prior DM-based methods \cite{graham2023denoising,liu2023unsupervised,heng2024out,Abdi_2025_ICCV}, as detailed in Appendix \ref{app:sec:dataset-baseline}.
For the OoD data, following the settings in \cite{gao2023diffguard}, four datasets are selected: Species \cite{he2024species196}, iNaturalist \cite{van2018inaturalist}, OpenImage-O \cite{wang2022vim} and ImageNet-O \cite{hendrycks2021natural}.

\noindent\textbf{Diffusion models and classifier-under-protection.} 
We utilize the diffusion denoising probabilistic model with checkpoints released by OpenAI\footnote{https://github.com/openai/guided-diffusion} \cite{dhariwal2021diffusion}.
The adopted checkpoint is solely pretrained on the ImageNet-1K training set, in line with the InD dataset setup.
In the main comparisons of Section \ref{sec:exp-main}, {\ours} uses the conditional $256\times256$ DM with the DDIM sampling \cite{song2020denoising}.
Other relevant settings are analyzed in Section \ref{sec:exp-diffusion}, including unconditional DMs and varied sampling time steps.
We do not consider another popular Latent Diffusion Models (LDMs) \cite{rombach2022high}, as the auto-encoder encoding the latent space of LDMs is trained on a much larger dataset than the ImageNet-1K. 
Hence, the InD for LDMs is substantially broader than ImageNet-1K, which is incompatible with our setup that adopts ImageNet-1K as the InD for fair comparisons.
For the classifier-under-protection, we employ the PyTorch-released \cite{paszke2019pytorch} checkpoints of ResNet50 \cite{he2016deep} and DenseNet121 \cite{huang2017densely} that are also pretrained on ImageNet-1K.

\begin{table*}[t]
    \centering
    \caption{Results of a variety of OoD detection methods on {ImageNet-1K} as InD data  with {ResNet50}. We highlight the {\bf best} and \underline{runner-up} results in bond font and with underlines, respectively.} 
    \label{tab:exp-imgnet-r50}
    \setlength{\tabcolsep}{1mm}
    \begin{tabular}{c|cc cc cc cc|cc}
    \toprule
    \multirow{3}{*}{method} & \multicolumn{8}{c|}{OoD data sets} & \multicolumn{2}{c}{\multirow{2}{*}{\bf AVERAGE}}\\
    & \multicolumn{2}{c}{Species} & \multicolumn{2}{c}{iNaturalist} & \multicolumn{2}{c}{OpenImage-O} & \multicolumn{2}{c|}{ImageNet-O} \\
    \cmidrule{2-11}
    & FPR$\downarrow$ & AUROC$\uparrow$ & FPR$\downarrow$ & AUROC$\uparrow$ &
    FPR$\downarrow$ & AUROC$\uparrow$ & FPR$\downarrow$ & AUROC$\uparrow$ & FPR$\downarrow$ & AUROC$\uparrow$ \\
    \midrule
    \multicolumn{11}{c}{\it non-diffusion-based}\\
    MSP \cite{hendrycks2016baseline} & 79.80 & 75.17 & 52.82 & 88.39 & 64.05 & 84.85 & 100.00 & 28.62 & 74.17 & 69.26 \\
    Energy \cite{liu2020energy} & 82.62 & 71.98 & 53.76 & 90.62 & 57.69 & 89.02 & 100.00 & 41.79 & 73.52 & 73.35 \\
    GradNorm \cite{huang2021importance} & 74.24 & 75.84 & 26.77 & 93.90 & 48.24 & 84.79 & 95.75 & 47.89 & 61.25 & 75.60 \\
    Mahalanobis \cite{lee2018simple} & 97.26 & 48.97 & 98.34 & 42.55 & 86.50 & 61.23 & 73.40 & 67.79 & 88.88 & 55.14 \\
    ReAct \cite{sun2021react} & 68.35 & 77.47 & 19.56 & 96.40 & 44.03 & 90.42 & 97.90 & 52.41 & 57.46 & 79.18 \\
    VRA \cite{xu2023vra} & 69.63 & 77.08 & {\bf 15.70} & {\bf97.14} & \underline{36.56} & {\bf92.94} & 95.55 & 60.79 & \underline{54.36} & \underline{81.99} \\
    KNN \cite{sun2022out} & 76.19 & 76.38 & 68.41 & 85.12 & 57.56 & 86.45 & 84.65 & 75.37 & 71.70 & 80.83 \\
    ViM \cite{wang2022vim} & 83.94 & 70.68 & 67.85 & 88.40 & 57.56 & 89.63 & 85.30 & 70.88 & 73.66 & 79.90 \\
    MLS \cite{hendrycks2022scaling} & 80.87 & 72.89 & 50.80 & 91.15 & 57.11 & 89.26 & 100.00 & 40.85 & 72.20 & 73.54 \\
    KPCA \cite{fang2024kernel} & 79.33 & 74.57 & 49.84 & 89.40 & 54.74 & 87.10 & {\bf60.95} & {\bf85.28} & 61.22 & 84.09 \\
    Mahala++ \cite{mullermahalanobis++} & 81.08 & 73.53 & 50.17 & 90.62 & 46.69 & 90.40 & \underline{64.95} & \underline{82.64} & 60.72 & 84.30 \\
    \midrule
    \multicolumn{11}{c}{\it diffusion-based}\\
    DDPM-OOD \cite{graham2023denoising} & 94.52 & 52.26 & 94.70 & 50.10 & 94.11 & 47.45 & 93.60 & 47.39 & 94.23 & 49.30 \\
    LMD \cite{liu2023unsupervised} & 96.13 & 55.66 & 94.65 & 60.04 & 93.50 & 51.36 & 92.35 & 50.82 & 94.16 & 54.47 \\
    DiffPath \cite{heng2024out} & 96.70 & 49.35 & 90.38 & 59.40 & 90.53 & 58.00 & 81.70 & 68.27 & 89.83 & 58.76 \\
    DiffPathV2 \cite{Abdi_2025_ICCV} & 94.54 & 43.92 & 95.57 & 45.30 & 95.35 & 49.81 & 84.45 & 62.32 & 92.48 & 50.34 \\
    DiffGuard \cite{gao2023diffguard} & 83.68 & 73.19 & 71.23 & 85.81 & 74.80 & 82.32 & 87.74 & 65.23 & 79.36 & 76.64 \\
    DiffGuard{\scriptsize+KNN} & 71.04 & 77.81 & 48.79 & 90.19 & 52.80 & 87.80 & 80.85 & 75.68 & 63.37 & 82.87 \\
    DiffGuard{\scriptsize+ViM} & 72.26 & 74.48 & 39.09 & 92.50 & 45.02 & 91.11 & 82.30 & 72.42 & 59.67 & 82.63 \\
    DiffGuard{\scriptsize+MLS} & 70.31 & 75.95 & 30.74 & 93.03 & 40.61 & 90.74 & 87.05 & 65.72 & 57.18 & 81.36 \\
    \cmidrule{2-11}
    \rowcolor{tabgray} {\ours} & {\bf 67.82} & {\bf 80.17} & \underline{17.96} & \underline{96.69}&	{\bf 35.02} & \underline{92.86} &	86.80 &	69.08 &	{\bf 51.90} & {\bf 84.70} \\
    \bottomrule
    \end{tabular}
\end{table*}

\noindent{\bf Evaluation metrics.}
Two metrics are calculated to evaluate the detection performance: False Positive Rate (FPR) of OoD samples at a 95\% True Positive Rate (TPR) on the InD test set, and Area Under the Receiver Operating Characteristic Curve (AUROC) w.r.t. varied pairs of FPR and TPR values.

\noindent\textbf{Baselines.} 
Two types of methods are considered as baselines in comparisons:
\begin{itemize}[leftmargin=*]
    \item {\it Non-diffusion-based}: These methods operate on the responses from the classifier-under-protection of the input $\bm x$.
    We select multiple strong detection baselines in experiments, covering logits-, features-, and gradients-based methods.
    Particularly, methods that leverage the Mahalanobis distance \cite{lee2018simple,mullermahalanobis++}, Energy function \cite{liu2020energy} and subspace projection \cite{fang2024kernel} are closely relevant with our {\ours}, and are included into comparisons.
    Refer to Appendix \ref{app:sec:related-work} for an outline on those involved baselines.
    \item {\it Diffusion-based}: We consider the first two paradigms reviewed in Sec.\ref{sec:related-work-ood-diffusion}, i.e., Generation-and-Comparison and Trajectory-Monitoring, due to their similar detection rationales with {\ours}, i.e., discrepancy evaluation on the DM's outputs or internal states, while the third Outlier-Synthesis-for-Training paradigm is skipped, since its mechanism, training classifier by augmented generative OoD outlier, fundamentally differs from {\ours}. 
    Comparisons cover several prevailing methods: DDPM-OOD \cite{graham2023denoising}, LMD \cite{liu2023unsupervised}, DiffGuard \cite{gao2023diffguard}, DiffPath \cite{heng2024out} and DiffPathV2 \cite{Abdi_2025_ICCV}.
\end{itemize}

\begin{table*}[t]
    \centering
    \caption{Results of a variety of OoD detection methods on {ImageNet-1K} with {DenseNet121}. The {\bf best} and \underline{runner-up} results are highlighted with bond fonts and underlines, respectively.}
    \label{tab:exp-imgnet-dn121}
    \setlength{\tabcolsep}{1mm}
    \begin{tabular}{c|cc cc cc cc|cc}
    \toprule
    \multirow{3}{*}{method} & \multicolumn{8}{c|}{OoD data sets} & \multicolumn{2}{c}{\multirow{2}{*}{\bf AVERAGE}}\\
    & \multicolumn{2}{c}{Species} & \multicolumn{2}{c}{iNaturalist} & \multicolumn{2}{c}{OpenImage-O} & \multicolumn{2}{c|}{ImageNet-O} \\
    \cmidrule{2-11}
    & FPR$\downarrow$ & AUROC$\uparrow$ & FPR$\downarrow$ & AUROC$\uparrow$ &
    FPR$\downarrow$ & AUROC$\uparrow$ & FPR$\downarrow$ & AUROC$\uparrow$ & FPR$\downarrow$ & AUROC$\uparrow$ \\
    \midrule
    \multicolumn{11}{c}{\it non-diffusion-based}\\
    MSP \cite{hendrycks2016baseline} & 80.59 & 74.74 & 49.60 & 89.04 & 67.17 & 83.72 & 98.60 & 47.34 & 73.99 & 73.71 \\
    Energy \cite{liu2020energy} & 80.38 & 71.65 & 39.77 & 92.66 & 56.77 & \underline{88.42} & 96.35 & 57.35 & 68.32 & 77.52 \\
    GradNorm \cite{huang2021importance} & 79.41 & 69.50 & \underline{26.78} & \underline{93.40} & 55.64 & 80.44 & 92.40 & 54.52 & 63.56 & 74.46 \\
    Mahalanobis \cite{lee2018simple} & 92.59 & 58.71 & 95.39 & 49.54 & 90.59 & 50.92 & 89.45 & 45.30 & 92.01 & 51.12 \\
    ReAct \cite{sun2021react} & 77.07 & 77.95 & 46.81 & 90.31 & 67.84 & 74.12 & 93.05 & 48.42 & 71.19 & 72.70 \\
    VRA \cite{xu2023vra} & \underline{75.88} & \underline{75.18} & 33.65 & 93.05 & \underline{55.29} & 82.63 & 84.90 & 63.80 & \underline{62.43} & 78.67 \\
    KNN \cite{sun2022out} & 89.82 & 68.76 & 77.22 & 82.30 & 70.85 & 81.49 & {\bf61.30} & {\bf85.13} & 74.80 & 79.42  \\
    MLS \cite{hendrycks2022scaling} & 79.86 & 72.56 & 39.73 & 92.80 & 58.09 & 88.41 & 97.50 & 56.31 & 68.79 & 77.52 \\
    KPCA \cite{fang2024kernel} & 87.81 & 69.58 & 69.76 & 83.12 & 69.40 & 80.30 & \underline{65.30} & \underline{83.11} & 73.07 & 79.03 \\
    Mahala++ \cite{mullermahalanobis++} & 88.00 & 67.04 & 56.04 & 89.04 & 55.85 & 85.95 & 72.25 & 78.28 & 68.04 & \underline{80.08} \\
    \midrule
    \multicolumn{11}{c}{\it diffusion-based}\\
    DDPM-OOD \cite{graham2023denoising} & 94.52 & 52.26 & 94.70 & 50.10 & 94.11 & 47.45 & 93.60 & 47.39 & 94.23 & 49.30 \\
    LMD \cite{liu2023unsupervised} & 96.13 & 55.66 & 94.65 & 60.04 & 93.50 & 51.36 & 92.35 & 50.82 & 94.16 & 54.47 \\
    DiffPath \cite{heng2024out} & 96.70 & 49.35 & 90.38 & 59.40 & 90.53 & 58.00 & 81.70 & 68.27 & 89.83 & 58.76 \\
    DiffPathV2 \cite{Abdi_2025_ICCV} & 94.54 & 43.92 & 95.57 & 45.30 & 95.35 & 49.81 & 84.45 & 62.32 & 92.48 & 50.34 \\
    DiffGuard \cite{gao2023diffguard} & 84.74 & 72.07 & 77.11 & 85.37 & 78.13 & 81.78 & 89.65 & 66.76 & 82.41 & 76.49 \\
    \cmidrule{2-11}
    \rowcolor{tabgray} {\ours} & {\bf74.76} & {\bf76.20} & {\bf22.86} & {\bf95.78} & {\bf44.79} & {\bf89.43} & 68.25 & 82.33 & {\bf52.67} & {\bf85.93} \\
    \bottomrule
    \end{tabular}
\end{table*}

\subsection{Main results}
\label{sec:exp-main}

Tables \ref{tab:exp-imgnet-r50} and \ref{tab:exp-imgnet-dn121} show detection FPR and AUROC results among a variety of non-diffusion-based and diffusion-based methods w.r.t. ResNet50 and DenseNet121 as the classifier-under-protection, respectively.
Note that DDPM-OOD, LMD, DiffPath and DiffPathV2 are agnostic of the classifier, and thereby exhibit the same detection results in Tables \ref{tab:exp-imgnet-r50} and \ref{tab:exp-imgnet-dn121}.
As neither the paper nor the code for DiffGuard provides sufficient implementation details of combining it with other methods, we are unable to report results of such combined variants for DiffGuard with DenseNet121 in Table \ref{tab:exp-imgnet-dn121}.

As shown in Tables \ref{tab:exp-imgnet-r50} and \ref{tab:exp-imgnet-dn121}, {\ours} outperforms those non-diffusion-based detection methods due to the application of the DM with superior distribution modeling capabilities.
Among those diffusion-based baselines, DDPM-OOD, LMD, DiffPath and DiffPathV2 show quite weak performance on this challenging ImageNet-1K InD dataset.
DDPM-OOD and LMD adopt perceptual distances to evaluate the discrepancy in raw images between $\bm x$ and its denoised or inpainted $\bm{\hat x}$.
It is noted that their effectiveness is largely confined to scenarios where the InD dataset exhibits pronounced and distinctive attributes, e.g., the human face dataset CelebA \cite{liu2015deep} as InD, as demonstrated in their original papers.
Regarding the complex natural images in ImageNet-1K, the perceptual metric fails to capture the discrepancy.
We provide several exemplar images for LMD in Appendix \ref{app:sec:visualization} to intuitively illustrate its ineffectiveness in natural images.
DiffPath and DiffPathV2 operate by computing simple statistics (e.g., norm, MSE, or SSIM) from the DM’s internal states and then fitting a classic kernel density model on these statistics.
While this pipeline may suffice for low-resolution images ($32\times32$ or $64\times64$) as demonstrated in their original papers, it fails to generalize to more complex, large-scale scenarios.
For $256\times256$ DMs on the large size ImageNet-1K in this InD setup, such simplistic statistics and shallow models cannot adequately capture the underlying discrepancy, leading to degraded performance.
DiffGuard introduces the classifier-under-protection into the DM's reverse process to guide generating semantically mismatched $\bm{\hat x}$ but resorts to perceptual distances, which leads to significantly improved performance over other DM-based methods.
Nevertheless, it still needs to be combined with some non-diffusion methods so as to achieve competitive results.
Within our {\ours} framework, the discrepancy from the DM is evaluated in deep representation spaces from the classifier with rich InD knowledge.
The devised metrics and associated representation refinement effectively capture the underlying discrepancy between $\bm x$ and $\bm{\hat x}$ in representation spaces, achieving superior detection performance over those non-diffusion-based and diffusion-based baselines.

\noindent{\bf Near- and far-OoD results.}
We further provide a fine-grained evaluation on near-OoD and far-OoD datasets.
This grouping is determined by image content and semantics:
Near-OoD images are similar to InD images, and far-OoD images deviate much from InD images in both semantic meaning and low-level statistics.
For ImageNet-1K as InD, the corresponding near-OoD includes NINCO \cite{bitterwolf2023ninco} and SSB-hard \cite{vaze2021open}, and far-OoD contains iNaturalist \cite{van2018inaturalist}, Texture \cite{cimpoi2014describing} and OpenImage-O \cite{wang2022vim}.
The average detection performance on these datasets is reported in Table \ref{tab:exp-imgnet-near-far-ood}, with full results provided in Table S4 of Appendix \ref{app:sec:complete-results}.
Note that we evaluate on all 50,000 images of the ImageNet-1K test set, unlike some prior works that use only 45,000 test images, which accounts for the possible difference in reported results.
Consistent with Tables \ref{tab:exp-imgnet-r50} and \ref{tab:exp-imgnet-dn121}, {\ours} maintains superior detection performance on both near-OoD and far-OoD images, demonstrating the effectiveness of its discrepancy evaluation framework.

\begin{table}[t]
    \centering
    \caption{Evaluation on near- and far-OoD on {ImageNet-1K} with {ResNet50}. The {\bf best} and \underline{runner-up} results are highlighted with bond fonts and underlines, respectively.}
    \label{tab:exp-imgnet-near-far-ood}
    \begin{tabular}{c|cc|cc}
    \toprule
    \multirow{2}{*}{Method} & \multicolumn{2}{c|}{Near-OoD (Avg.)} & \multicolumn{2}{c}{Far-OoD (Avg.)} \\
    \cmidrule{2-5}
    & FPR$\downarrow$ & AUROC$\uparrow$ & FPR$\downarrow$ & AUROC$\uparrow$ \\
    \midrule
    \multicolumn{5}{c}{\it non-diffusion-based}\\
    MSP \cite{hendrycks2016baseline} & 80.70 & 76.01 & 61.05 & 84.56 \\
    Energy \cite{liu2020energy} & 81.02 & 75.89 & 54.64 & 88.79 \\
    GradNorm \cite{huang2021importance} & 77.55 & 73.01 & 35.95 & 89.77 \\
    ReAct \cite{sun2021react} & 75.19 & 76.40 & 36.66 & 92.38 \\
    VRA \cite{xu2023vra} & 77.73 & 75.79 & \underline{24.65} & \underline{95.24} \\
    KNN \cite{sun2022out} & 86.15 & 65.16 & 43.46 & 89.66 \\
    NNGuide \cite{park2023nearest} & \underline{74.98} & \underline{77.32} & 31.20 & 93.10 \\
    MLS \cite{hendrycks2022scaling} & 80.52 & 76.45 & 54.57 & 88.89 \\
    KPCA \cite{fang2024kernel} & 83.58 & 69.16 & 37.85 & 91.55 \\
    Mahala++ \cite{mullermahalanobis++} & 82.74 & 72.09 & 34.20 & 93.27 \\
    \midrule
    \multicolumn{5}{c}{\it diffusion-based}\\
    DDPM-OOD \cite{graham2023denoising} & 93.54 & 50.68 & 94.47 & 45.38 \\
    LMD \cite{liu2023unsupervised} & 94.42 & 51.88 & 89.46 & 57.25 \\
    DiffPath \cite{heng2024out} & 92.18 & 52.63 & 86.45 & 62.99 \\
    DiffPathV2 \cite{Abdi_2025_ICCV} & 91.56 & 52.70 & 93.20 & 51.09 \\
    DiffGuard \cite{gao2023diffguard} & 83.63 & 70.65 & 67.18 & 84.51 \\
    \cmidrule{2-5}
    \rowcolor{tabgray} {\ours} & {\bf74.03} & {\bf78.60} & {\bf20.87} & {\bf95.84} \\
    \bottomrule
    \end{tabular}
\end{table}

\subsection{Ablation studies}
\label{sec:exp-ablation}

\noindent\textbf{Individual effect of $\epsilon_{\rm feat}$ and $\epsilon_{\rm logt}$.}
In this part, two sets of evaluations are conducted. 
Firstly, we investigate the individual effectiveness of our devised metrics $\epsilon_{\rm feat}$, $ \epsilon_{\rm logt}$ and their ensemble. 
Secondly, regarding the Mahalanobis distance in $\epsilon_{\rm feat}$, we demonstrate the significance of the training feature covariance matrix ${\bf\Sigma}_{\bm h}$ by considering a well-recognized $\ell_2$-normalized $\ell_2$ distance ${\ell_2}({\bm h}_{\bm {x}},{\bm h}_{\bm {\hat x}})=\|\frac{{\bm h}_{\bm {x}}}{\|{\bm h}_{\bm {x}}\|_2}-\frac{{\bm h}_{\bm {\hat x}}}{\|{\bm h}_{\bm {\hat x}}\|_2}\|_2$ \cite{sun2022out}, since the Mahalanobis distance without the covariance degenerates to the $\ell_2$ distance.
Another common cosine similarity distance ${\rm cos}({\bm h}_{\bm {x}},{\bm h}_{\bm {\hat x}})=\frac
{\bm {h}_{\bm {\hat x}}^\top\bm {h}_{\bm {x}}}
{\|\bm {h}_{\bm {\hat x}}\|_2\cdot\|\bm {h}_{\bm {x}}\|_2}$ is also incorporated into comparisons.
Table \ref{tab:ablation-prob-feat-short} shows the individual effects of our devised metrics and their alternatives with results averaged over 4 OoD datasets (Species, iNaturalist, OpenImage-O and ImageNet-O).
Full results on each OoD dataset are provided in Appendix \ref{app:sec:complete-results}.
We can have the following observations.
\begin{itemize}[leftmargin=*]
    \item It shows that the individual metric in either feature space or logit space gives distinctively inferior detection performance than their ensemble.
    Such results verify the significance of fully investigating both covariate shift in features and concept shift in logits
    from the classifier in assessing the discrepancy between $\bm x$ and $\bm{\hat x}$.
    Particularly, in Appendix \ref{app:sec:visualization}, we provide several exemplar images that are detected by exactly one of the two metrics $\epsilon_{\text{feat}}$ and $\epsilon_{\text{logit}}$, revealing their complementary role.
    \item In the feature space, both the $\ell_2$ distance and the cosine similarity show weaker detection performance than the Mahalanobis distance.
    This implies that each dimension in the features contributes differently to the discrepancy measurement, which should not be overlooked.
    The Mahalanobis distance in {\ours} introduces the training feature covariance matrix ${\bf\Sigma}_{\bm h}$ to assign corresponding weights to dimensions of different variances, thereby leading to better detection results.
    Besides, the $\ell_2$ distance and the cosine similarity exhibit the same detection value.
    The reasons behind can be that the different values in the two feature distances do not change the order of the detection scores of samples in InD and OoD datasets, which brings the same FPR and AUROC results.
\end{itemize}

\begin{table}[t]
    \centering
    \caption{The individual effects of different alternatives of the metrics in the feature space and logit space of {\ours}.}
    \label{tab:ablation-prob-feat-short}
    \begin{tabular}{c|cc}
    \toprule
    \multirow{2}{*}{Score} & \multicolumn{2}{c}{Detection results (Avg.)} \\
    \cmidrule{2-3}
    & FPR$\downarrow$ & AUROC$\uparrow$ \\
    \midrule
    $\epsilon_{\rm feat}$ & 63.96 & 80.79 \\
    $\epsilon_{\rm logt}$ & 77.12 & 71.80 \\
    Ensemble & {\bf51.90} & {\bf84.70} \\
    \midrule
    $\epsilon_{\rm feat}$ ($\ell_2$) & 69.35 & 78.48 \\
    $\epsilon_{\rm feat}$ (cosine) & 69.35 & 78.48 \\
    \bottomrule
    \end{tabular}
\end{table}

\begin{table}[t]
    \centering
    \caption{Ablation studies on the subspace-based representation refinement strategy of {\ours}.}
    \label{tab:ablation-boosting-short}
    \begin{tabular}{c|cc}
    \toprule
    \multirow{2}{*}{Implementation} & \multicolumn{2}{c}{Detection results (Avg.)}\\
    \cmidrule{2-3}
    & FPR$\downarrow$ & AUROC$\uparrow$ \\
    \midrule
    w/o refine & 62.09 & 80.51 \\
    Refine on ${\bm h}_{\bm{x}},{\bm z}_{\bm{x}},{\bm h}_{\bm{\hat x}},{\bm z}_{\bm{\hat x}}$ & 64.83 & 79.87 \\
    Refine on ${\bm h}_{\bm{x}},{\bm z}_{\bm{x}}$ & 82.58 & 67.43 \\
    Refine on ${\bm h}_{\bm{\hat x}},{\bm z}_{\bm{\hat x}}$ & {\bf51.90} & {\bf84.70} \\
    \bottomrule
    \end{tabular}
\end{table}

\noindent\textbf{Effect of subspace-based representation refinement.}
Ablation studies on the implementation position of our subspace-based representation refinement strategy are executed.
Table \ref{tab:ablation-boosting-short} shows the average results over the selected 4 OoD datasets; the full results are provided in Appendix \ref{app:sec:complete-results}.
Our {\ours} refines the representations of the DM-generated ${\bm{\hat x}}$ (``refine on ${\bm h}_{\bm{\hat x}},{\bm z}_{\bm{\hat x}}$'').
Aside from this setup, three alternative implementations are exploited in the ablations, including {\ours} without the refinement strategy (``w/o refine''), {\ours} with refinement on representations of $\bm {x}$ (``refine on ${\bm h}_{\bm{x}},{\bm z}_{\bm{x}}$'') and {\ours} with refinement on representations of both $\bm {x}$ and $\bm {\hat x}$ (``refine on ${\bm h}_{\bm{x}},{\bm z}_{\bm{x}},{\bm h}_{\bm{\hat x}},{\bm z}_{\bm{\hat x}}$'').
As demonstrated in Table \ref{tab:ablation-boosting-short}, we have the following observations and discussions.
\begin{itemize}[leftmargin=*]
\item The representation refinement significantly advocates the discrepancy assessment of {\ours}, validated by the substantial detection performance improvements on top of {\ours} without refinement (``w/o refine'').
This is because that the subspace projection removes the abnormal information in representations of the DM-generated $\bm{\hat x}$, and enhances their alignment with InD, thereby rendering the discrepancy in representations between $\bm x$ and $\bm{\hat x}$ more pronounced.
\item In contrast, the refinement strategy executed on representations of $\bm{x}$ or of both $\bm{x}$ and $\bm{\hat x}$ shows a negative impact on the detection performance. 
The reasons behind lie in that it remains unclear whether the input image $\bm x$ is from InD or OoD. Thereby, we cannot simply remove the anomalous information in $\bm x$'s representations by subspace projection, which otherwise would be detrimental to a correct assessment on the discrepancy in representations between $\bm x$ and $\bm{\hat x}$.
\end{itemize}

\begin{figure}[t]
    \centering
    \includegraphics[width=0.85\linewidth]{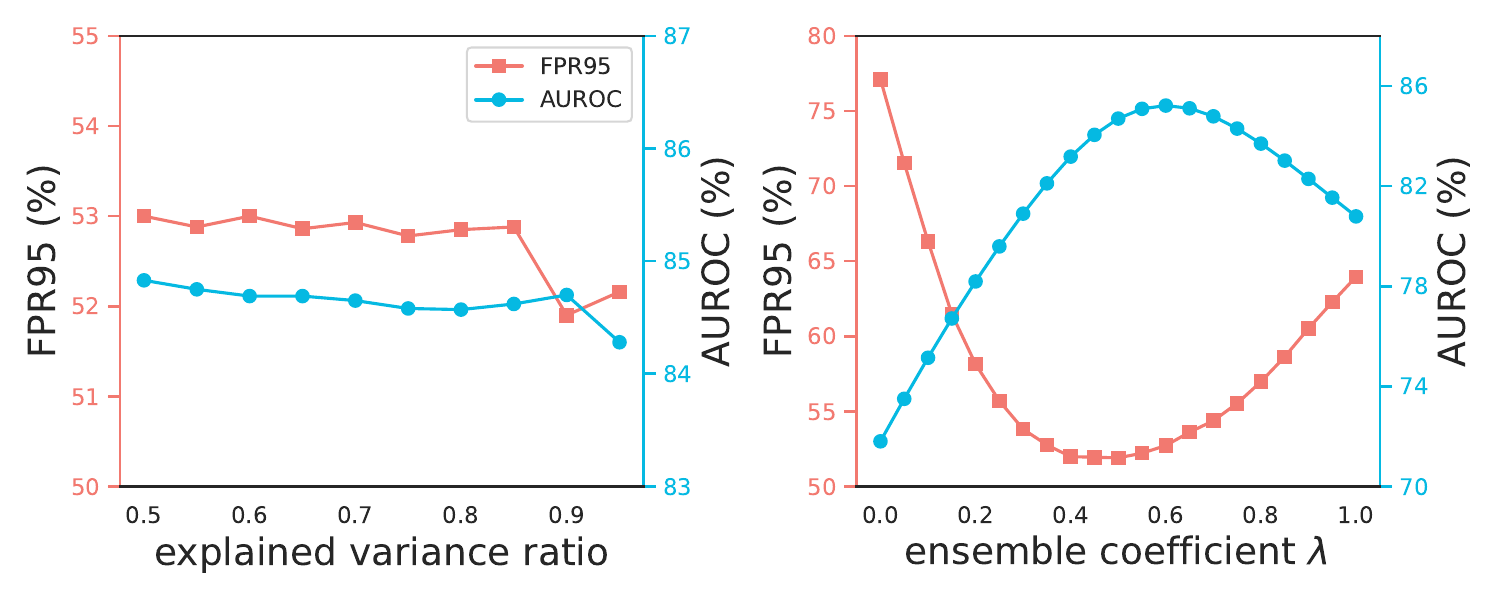}
    \caption{{\bf Left}: Detection results w.r.t varied values of explained variance ratio. {\bf Right}: Detection results w.r.t. varied ensemble coefficients $\lambda$. The average detection FPR and AUROC results over the 4 OoD datasets are reported.}
    \label{fig:exp-sensitivity}
\end{figure}

\subsection{Sensitivity analysis}
\label{sec:exp-sensitivity}

\noindent\textbf{Effect of subspace dimension.} The subspace-based representation refinement strategy in {\ours} learns projection matrices ${\bf U}_{\bm h}\in\mathbb{R}^{m\times {q_h}}$ and  ${\bf U}_{\bm z}\in\mathbb{R}^{c\times {q_z}}$ from the training representations by preserving most of the principal information of the InD training data.
Here, the amount of preserved principal information in the projected subspace directly determines the subspace dimensions $q_h$ and $q_z$, and affects the effectiveness of refinement, and is measured by the {\it explained variance ratio}, i.e., the ratio of the sum of preserved top eigenvalues over the sum of all eigenvalues.
A sensitivity analysis on explained variance ratio is provided to investigate the effect of the amount of the preserved information in the projected subspace, with detection results shown in the left panel of Figure \ref{fig:exp-sensitivity}.
As illustrated, a smaller explained variance ratio implies less InD information preserved in the subspace and less refinement through the subspace projection, which thereby leads to weaker detection performance of {\ours}. 
Through our extensive experiments,  we suggest to retain 90\% of the explained variance by default for {\ours}.

\noindent\textbf{Effect of ensemble coefficient $\lambda$.}
In our ensemble scheme, the coefficient $\lambda$ in Eqn.\eqref{eq:ddos-final} balances the two metrics $\epsilon_{\rm feat}$ and $\epsilon_{\rm logt}$ on the feature and logit representations, respectively.
A sensitivity analysis on varied values of $\lambda$ is conducted to investigate the effect of ensembling such two metrics, with results shown in the right panel of Figure \ref{fig:exp-sensitivity}.
It can be observed from Figure \ref{fig:exp-sensitivity} that a rather balanced coefficient $\lambda=0.5$ brings the lowest detection FPR value and also the highest AUROC value of {\ours} and  thus is thereby suggested for practitioners with our {\ours} method. 
This analysis also indicates that the discrepancy from the covariate shift and concept shift is nearly equally important in contributing to the superior detection performance, further verifying their indispensable significance.

\begin{figure}[t]
    \centering
    \includegraphics[width=0.85\linewidth]{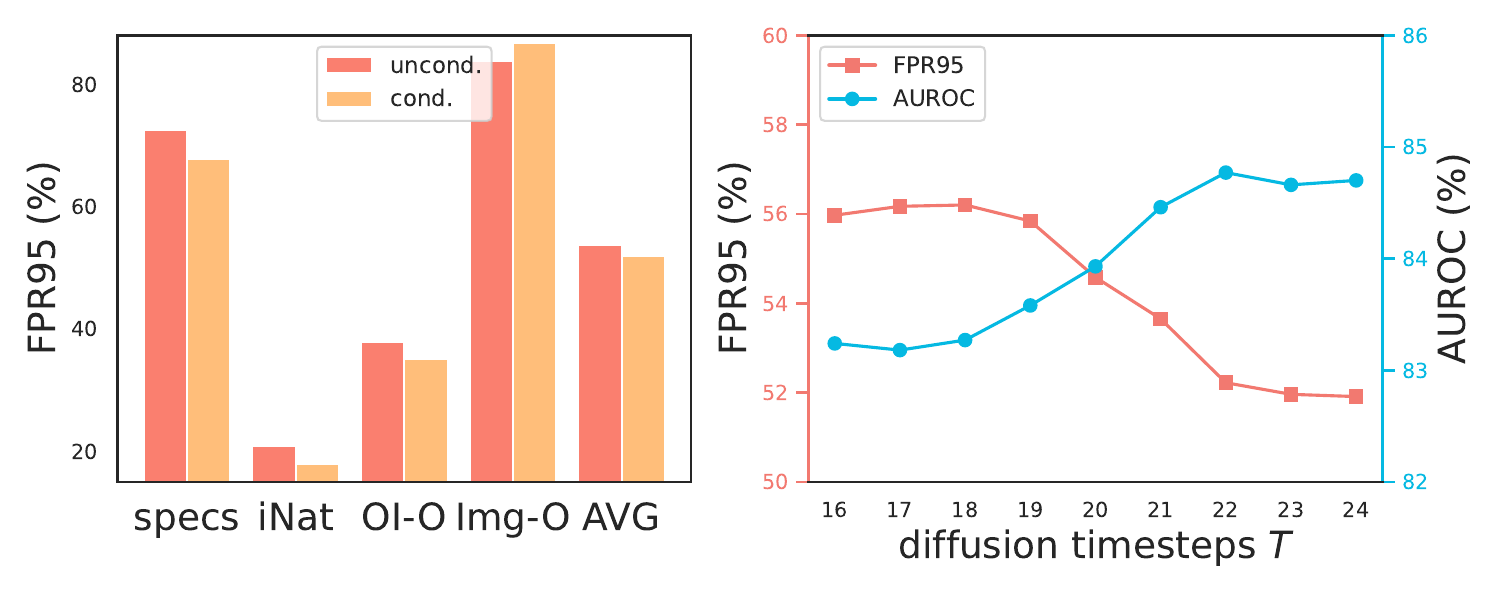}
    \caption{{\bf Left}: Comparisons on the detection performance w.r.t. conditional (cond.) and unconditional (uncond.) DMs in {\ours} on each OoD dataset. {\bf Right}: Detection results w.r.t. varied diffusion time steps $T$ of {\ours} in generating $\bm{\hat x}$. The average detection FPR and AUROC results of {\ours} over the 4 OoD datasets are reported.}
    \label{fig:exp-dm}
\end{figure}

\subsection{Further evaluations on diffusion models}
\label{sec:exp-diffusion}

\noindent\textbf{Conditional v.s. unconditional DMs.}
{\ours} adopts a conditional DM to generate $\bm{\hat x}$ \cite{dhariwal2021diffusion}. 
That is, during the reverse process of the DM, the predictive labels from the classifier-under-protection $f$ on the input $\bm x$ are leveraged as a conditional signal to guide the generation of $\bm{\hat x}$.
This condition mechanism benefits the quality of the generated images and forces $\bm{\hat x}$ to be more aligned with InD, thereby promoting OoD detection.
To validate the importance of this conditional generation, an ablation study is conducted by replacing the conditional DM with an unconditional one, where $\bm {\hat x}$ is recovered without any label information.
The results in the left panel of Figure \ref{fig:exp-dm} demonstrate the inferior performances of unconditional DMs than conditional DMs for {\ours}. 
The DM-generated $\bm {\hat x}$ with the label guidance approaches more closely towards the InD than those without the label guidance, thus contributing to a more significant discrepancy between $\bm {x}$ and $\bm {\hat x}$ and improved detection performance.

\noindent\textbf{Effect of varied diffusion time steps.}
{\ours} adopts the DDIM sampling \cite{song2020denoising} for fewer sampling steps $T$ in the reverse processes of the DM.
Different steps $T$ significantly affect the quality of generated images $\bm {\hat x}$, and thereby influence the detection performance.
A wide range of the reverse time steps $T$ are exploited for a sensitivity analysis on the detection results of {\ours}, shown the right panel of Figure \ref{fig:exp-dm}.

As the diffusion time step $T$ increases, the quality of the generated image $\bm{\hat x}$ improves and its alignment with InD strengthens.
This enhancement makes the discrepancy between $\bm x$ and $\bm{\hat x}$ more pronounced.
Consequently, as validated by the results in Figure \ref{fig:exp-dm}, OoD detection performances improve with larger values of $T$, implying the importance of high-quality generative images for our {\ours} framework.
We also visualize some examples w.r.t. different time steps $T$ in Appendix \ref{app:sec:visualization} for an intuitive demonstration.
In the main comparisons of Sec.\ref{sec:exp-main}, {\ours} adopts a default value of $T=24$ following the recommended settings in \cite{dhariwal2021diffusion}.

\section{Conclusion and discussion}
\label{sec:conclusion}
In this work, we propose {\ours}, a framework that advances DM-based OoD detection by reframing the core problem of discrepancy assessment.
Our key insight is that the underlying discrepancy between an input $\bm x$ and its DM-generated counterpart $\bm{\hat x}$  is most pronounced within representation spaces with rich InD knowledge, rather than being captured by human-perceptual metrics of visual similarity.
To this end, we leverage the representation spaces from the classifier-under-protection that is trained on InD data.
Within such spaces, representations of $\bm{\hat x}$, generated to align with InD, exhibit discernible discrepancy with the representations of the inputs $\bm{x}$, which can be characterized for OoD detection, as empirically demonstrated in our work.
Building on such spaces, {\ours} introduces tailored metrics to quantify two complementary aspects of the underlying discrepancy between $\bm x$ and $\bm{\hat x}$, i.e., the Mahalanobis distance in the feature space capturing geometric misalignment (covariate shift) and the Energy function in the logit space revealing semantic uncertainty (concept shift).
{\ours} further integrates a subspace-based strategy to refine representations of $\bm{\hat x}$ by projecting them to an InD subspace, enhancing their alignment with InD.
Extensive experiments validate the superior detection performances of {\ours} over a wide range of diffusion-based and non-diffusion-based detection methods.

The limitation of this work can lie in the computation costs in generating the synthesis $\bm{\hat x}$ via the diffusion model, which is though also a common limitation of DM-based methods.
We provide a detailed analysis into the computational overhead of {\ours} in Appendix \ref{app:sec:compute-overhead}.
One could mitigate this issue by introducing advanced acceleration techniques for diffusion models, such as lightweight model checkpoints and faster sampling schedules.

\bibliographystyle{unsrt}  
\bibliography{reference}

\clearpage

\setcounter{table}{0}
\renewcommand*{\thetable}{S\arabic{table}}
\setcounter{figure}{0}
\renewcommand*{\thefigure}{S\arabic{figure}}

\begin{appendices}

This appendix provides supplementary materials for the main paper, including related work, implementation details, datasets, computational cost analysis, visualizations, and complete experimental results.
\begin{itemize}
\item Appendix \ref{app:sec:related-work} provides a supplementary review of non-diffusion-based OoD detection methods and briefly explains their principles and differences from {\ours}.
\item Appendix \ref{app:sec:implementation} describes the two-phase implementation of DDR, including image preprocessing, representation preprocessing, and min-max normalization for ensembling.
\item Appendix \ref{app:sec:dataset-baseline} lists the OoD datasets used in experiments and notes that existing diffusion-based OoD methods have not been thoroughly explored on large-scale ImageNet-1K under standard OoD detection setups.
\item Appendix \ref{app:sec:compute-overhead} analyzes the computation overhead in the offline and online stages of {\ours}.
\item Appendix \ref{app:sec:visualization} shows failure cases of LMD, visual comparisons of generated images with different diffusion timesteps, and examples where the two metrics of {\ours} detect different types of OoD samples.
\item Appendix \ref{app:sec:complete-results} presents full detection results corresponding to the main text, including detailed performance metrics on each OoD dataset.
\end{itemize}

\section{Related work}
\label{app:sec:related-work}

Diffusion models \cite{sohl2015deep,ho2020denoising,song2020score,song2020denoising,dhariwal2021diffusion}, due to their powerful capability in generating high-fidelity images, have brought a fresh generative perspective to OoD detection \cite{graham2023denoising,liu2023unsupervised,gao2023diffguard,heng2024out,Abdi_2025_ICCV,yang2024diffusion,du2024dream,pmlr-v267-liao25g,yoon2025diffusion}, as thoroughly reviewed in Sec.III of the main text.
In this section, we supplement more related work on another prevalent series of non-DM-based detection approaches.

Mainstream detection methods solely leverage a discriminative model, typically the classifier-under-protection, a DNN model trained on InD data, and utilize its responses to characterize the InD-OoD discrepancy, without the incorporation of a generative model.
Such network responses can be the predictive logits \cite{hendrycks2016baseline,liu2020energy,hendrycks2022scaling}, the network-parameter gradients \cite{huang2021importance}, the network-layer features \cite{lee2018simple,sun2021react,xu2023vra,sun2022out,fang2024kernel,mullermahalanobis++,park2023nearest} and their ensemble \cite{fang2024revisiting}.
Proper metrics are then adopted on the tackled responses to quantify the likelihood of $\bm {x}$ being from the InD, such as the Energy function on logits \cite{liu2020energy}, appropriate distances \cite{sun2022out,mullermahalanobis++} and reconstruction errors \cite{fang2024kernel} on features, or the norms on parameter gradients \cite{huang2021importance}.
These methods have provided beneficial insights and practical implementation that significantly advance understandings of the InD-OoD discrepancy.

In our experiments, a variety of non-DM-based approaches are incorporated as strong baselines, covering the aforementioned different aspects.
We give a brief outline on these methods below.

\paragraph{Logits-based and Gradients-based}
MSP \cite{hendrycks2016baseline}, Energy \cite{liu2020energy}, and MLS \cite{hendrycks2022scaling} work on the predictive logits from the classifier-under-protection to distinguish InD and OoD data.
MLS simply takes the largest logit value as the detection score, while MSP chooses the largest softmax value.
Energy adopts the energy function on logits for OoD detection.
GradNorm \cite{huang2021importance} calculates the norm of parameter gradients w.r.t. a KL divergence loss between logits and a uniform distribution.
The essential rationale behind these methods lies in that InD samples tend to produce logits with a peak value on a class under the InD-trained classifier-under-protection, while OoD samples are likely to yield uniformly-distributed logits.

\paragraph{Feature-distance-based}
KNN \cite{sun2022out} and NNGuide \cite{park2023nearest} apply nearest neighbor searching on the training features.
KNN targets the $\ell_2$ distance, while NNGuide improves this by incorporating the Energy score.
Mahalanobis \cite{lee2018simple} and Mahalanobis++ \cite{mullermahalanobis++}, as their names suggest, leverage the Mahalanobis distance on features for OoD detection in different ways.
KPCA \cite{fang2024kernel} identifies a non-linear feature subspace where the InD-OoD discrepancy gets promoted, and leverages the reconstruction error as the distance for OoD detection.

\paragraph{Feature-shaping-based}
Such methods modify the features to remove those anomaly information causing OoD predictions, thereby enhancing OoD detection.
Their detection score is ultimately defined in a logits-based way, usually by applying the Energy function on logits w.r.t. the modified features.
A pioneering work ReAct \cite{sun2021react} adopts a simple feature-clipping way by clipping those extremely large feature values, then VRA \cite{xu2023vra} improves ReAct by a fine-grained clipping technique.

Another relevant work SISOM \cite{schmidtunified} utilizes a combination of Energy scores and quotients for OoD detection, which appears a similar form as the Energy ratio in {\ours}.
The differences between SISOM and {\ours} are also significant: SISOM does not incorporate a diffusion model and solves an optimization problem under the framework of active learning for OoD detection.
As the code of SISOM is not available, we fail to reproduce its results, and thereby omit comparisons with SISOM.

\section{Implementation of {\ours}}
\label{app:sec:implementation}

{\ours} can be decomposed into two phases: the generation phase where the DM is employed to generate a counterpart $\bm{\hat x}$ from the input $\bm x$ (cf. Eqns.\eqref{eq:dm-forward} and \eqref{eq:dm-reverse}), and the evaluation phase where discrepancy assessment is conducted in the representation space of the classifier-under-protection given $\bm x$ and $\bm{\hat x}$ for OoD detection.
We report some important implementation details of {\ours}, which can also be found in our released code.

\noindent{\bf Image preprocessing.}
In the generation phase, before the forward diffusion process of the DM, the input images $\bm x$ are resized and cropped to $256\times256\times3$ and normalized to $[-1,1]$, following the requirements of the released checkpoints of diffusion models in \cite{dhariwal2021diffusion}.
In the evaluation phase, before the forward propagation of the classifier-under-protection, input images $\bm x$ and their DM-generated counterparts $\bm{\hat x}$ are resized to $224\times224\times3$ with the mean-variance normalization, as required by the PyTorch-released checkpoints of the classifier-under-protection.

\noindent{\bf Representation preprocessing.}
In the evaluation phase, for features ${\bm h}_{\bm{\hat x}}$ of the DM-generated $\bm{\hat x}$, some widely-recognized preprocessings on $\bm{\hat h}$ are employed to facilitate the subspace projection.
To be specific, the extremely high feature values in ${\bm h}_{\bm{\hat x}}$ are clipped \cite{sun2021react}, and an $\ell_2$ normalization is conducted on the clipped features \cite{sun2022out,fang2024kernel}.
Such standard preprocessings bring features ${\bm h}_{\bm{\hat x}}$ to the same scale, and ensure numerical stability for the subsequent subspace projection.
The logits ${\bm z}_{\bm{\hat x}}$ of $\bm{\hat x}$ are attained w.r.t. the clipped features.
Note that such preprocessings are not applied to representations ${\bm h}_{\bm x}$ and ${\bm z}_{\bm x}$ of the inputs $\bm x$ in the evaluation phase, since the subspace projection is not executed on ${\bm h}_{\bm x}$ and ${\bm z}_{\bm x}$.
Detailed implementations can be found in our released code.

\noindent{\bf Ensemble preprocessing.}
The Mahalanobis distance on features $\epsilon_{\rm feat}$ and the Energy ratio on logits $\epsilon_{\rm logt}$ are of different magnitudes, thereby a normalization preprocessing before their ensemble is essential for a meaningful and stable fusion.
{\ours} adopts the min-max normalization based on statistics from the available InD test set.

\section{Datasets and baselines}
\label{app:sec:dataset-baseline}

In our comparisons, the selected 4 OoD datasets in Table I and Table II are outlined as follows.
\begin{itemize}[leftmargin=*]
    \item Species \cite{he2024species196} includes both labeled and unlabeled images of invasive species with bounding box annotations. 
    10,000 images from Species are sampled by OpenOOD \cite{yang2022openood} for the OoD detection task.
    \item iNaturalist \cite{van2018inaturalist} contains natural fine-grained images of different species of plants and animals. 
    10,000 images are sampled from the selected concepts by OpenOOD \cite{yang2022openood} for OoD detection.
    iNaturalist is also one of the far-OoD datasets w.r.t. ImageNet-1K as InD.
    \item OpenImage-O \cite{wang2022vim} is collected image-by-image from the test set of a large image dataset OpenImage-V3, and contains 
    17,632 images.
    OpenImage-O is also one of the far-OoD datasets w.r.t. ImageNet-1K as InD.
    \item ImageNet-O \cite{hendrycks2021natural} is gathered from the ImageNet-22K dataset with ImageNet-1K classes deleted. 
    The 2,000 images in ImageNet-O are even adversarially filtered, and are difficult to be distinguished from the InD ImageNet-1K dataset.
\end{itemize}

Besides, the near-OoD and far-OoD datasets w.r.t. ImageNet-1K as InD include the following 3 additional datasets:
\begin{itemize}[leftmargin=*]
    \item NINCO \cite{bitterwolf2023ninco} is a carefully-curated near-OoD dataset particularly for ImageNet-1K as InD.
    In NINCO, all 5,879 images across 64 classes are manually verified to contain no objects from any of the 1,000 ImageNet-1K classes, eliminating the contamination issue that plagued earlier datasets. 
    \item SSB-hard \cite{vaze2021open} is the other near-OoD dataset particularly for ImageNet-1K as InD.
    It is constructed by selecting a subset of ImageNet-21K samples that are semantically close to ImageNet-1K classes, making them particularly difficult to distinguish from InD data.
    \item Texture \cite{cimpoi2014describing} covers various types of surface texture with 5,640 images of 47 categories, and is adopted as the far-OoD dataset.
\end{itemize}

We elaborate the setups of InD and OoD datasets in the selected DM-based OoD detection baselines \cite{graham2023denoising,liu2023unsupervised,gao2023diffguard,heng2024out,Abdi_2025_ICCV} to highlight that the large-scale, complex ImageNet-1K dataset as a standard InD setting remains underexplored.

Most of the experiments in \cite{graham2023denoising,liu2023unsupervised,heng2024out,Abdi_2025_ICCV} are conducted on small-scale images ($32\times32$) with relatively simple content, such as CIFAR10 \cite{krizhevsky2009learning}, MNIST \cite{lecun2002gradient} and SVHN \cite{netzer2011reading}. 
While diffusion models on larger image sizes are considered in \cite{graham2023denoising,liu2023unsupervised}, e.g., $128\times128$ in \cite{graham2023denoising} and $256\times256$ in \cite{liu2023unsupervised}, the InD training datasets are highly specialized, such as the medical images in \cite{graham2023denoising} and the human face images in \cite{liu2023unsupervised}, rather than diverse, natural image collections.
Although ImageNet-1K has been used in \cite{heng2024out,Abdi_2025_ICCV}, the image size is limited to only $64\times64$. 
Moreover, these studies \cite{heng2024out,Abdi_2025_ICCV} apply DMs trained on ImageNet-1K to distinguish a different InD dataset from OoD data, e.g., differentiating CIFAR10 from SVHN, rather than separating ImageNet-1K from other OoD datasets, which deviates from a strict OoD detection setup in InD and OoD datasets.
DiffGuard \cite{gao2023diffguard} leverages diffusion models trained on the full-size ImageNet-1K and adopts correct setups for InD and OoD datasets. Our empirical settings thereby basically follow that in \cite{gao2023diffguard}.

\section{Computational overhead}
\label{app:sec:compute-overhead}

{\ours} can be divided into an offline stage and an online stage.
We analyze the computation cost of {\ours} in the two stages, respectively.

In its offline stage, {\ours} computes {\it (i)} the training feature covariance matrix ${\bf\Sigma}_{\bm h}$, {\it (ii)} the mean training features and logits ${\bm\mu}_{\bm h}$ and ${\bm\mu}_{\bm z}$, and {\it (iii)} the projection matrices ${\bf U}_{\bm h}$ and ${\bf U}_{\bm z}$, which will be leveraged to calculate the detection score in the later online stage.
These are obtained by performing a single forward pass of the entire training dataset through the classifier-under-protection, followed by PCA on the corresponding training features and logits.
To be specific, in experiments on ImageNet-1K, this forward pass over 1,281,167 training images using ResNet50 takes approximately 760 seconds on an NVIDIA GTX 4090 GPU.
Computing the $2,048\times2,048$ feature covariance matrix and the $1,000\times 1,000$ logit covariance matrix from the corresponding $1,281,167\times2,048$ and $1,281,167\times1,000$ training features and logits then takes only around 30 seconds on CPU.
The subsequent eigen-decomposition to the two covariance matrices requires less than 0.5 seconds also on CPU.
Overall, the offline stage of {\ours} incurs only modest computational overhead and, importantly, is performed entirely in advance, thus adding no extra cost to online detection.

In its online stage, the computational overhead of {\ours} primarily arises from the forward diffusion process (adding noise to the input image as in Eqn.\eqref{eq:dm-forward} of the main text), and the reverse diffusion process (denoising via the diffusion model as in Eqn.\eqref{eq:dm-reverse} of the main text).
In experiments on ImageNet-1K, running the two diffusion processes on a $256\times256$ image using the checkpoint\footnote{https://github.com/openai/guided-diffusion} with 1,000 diffusion steps takes around 2.3 seconds on an NVIDIA GTX 4090 GPU.
Thus, {\ours} introduces a 2.3-second latency from the involved diffusion model.
We note that this latency can be further reduced by adopting latest advances in diffusion models, such as faster sampling strategies.
Despite this, {\ours} offers distinct advantages of strong detection performance and deeper insight into the distributional discrepancy between samples.

\section{Visualization}
\label{app:sec:visualization}
\paragraph{Failure cases of LMD}
We provide several examples in Table \ref{app:tab:ldm-results} to intuitively demonstrate the ineffectiveness of LMD \cite{liu2023unsupervised} in our empirical setup.
As shown in Table \ref{app:tab:ldm-results}, given a DM pretrained on the ImageNet-1K with complex and natural images, the inpainted OoD images can also be successfully recovered with minor discrepancy with the original OoD images.
The adopted perceptual metrics in LMD thereby cannot well distinguish such discrepancy for InD and OoD input images, leading to the weak detection performance of LMD in our experiments.
LMD is demonstrated to be mostly effective under the scenario that the InD dataset is highly specialized, e.g., the human face dataset, as reported in its original paper.

\paragraph{Generated images w.r.t. varied diffusion time steps}
Table \ref{app:tab:ddr-step} demonstrates the visualization of the generated images w.r.t. different time steps $T$ in the reverse diffusion process.
We can observe that a smaller number of $T$ cannot generate high-quality images, thereby hurting the detection performance, as illustrated in the right panel of Figure \ref{fig:exp-dm}.
Moreover, one can also find that the underlying discrepancy between $\bm x$ and $\bm{\hat x}$ cannot be simply captured via perceptual metrics on their visual similarity, validating the perspective of our detection framework {\ours}.

\paragraph{Complementary cases under two metrics of {\ours}}
Table~\ref{app:tab:complementary-samples} presents representative OoD samples from OpenImage-O that are detected by exactly one of the two proposed metrics of {\ours}, revealing the complementary role of the feature-distance score $\epsilon_{\text{feat}}$ and the logit-energy ratio $\epsilon_{\text{logit}}$.

In Case~A, the inputs are text-heavy document images whose typographic layout superficially resembles certain ImageNet categories (e.g., book covers), causing the logit-energy ratio to remain above the detection threshold.
The diffusion model, however, reconstructs them as natural photographic scenes, producing a large feature-space discrepancy that $\epsilon_{\text{feat}}$ successfully captures.
In Case~B, the inputs are natural photographs that closely match the low-level statistics of ImageNet, so $\epsilon_{\text{feat}}$ cannot distinguish them from InD.
Yet the diffusion process maps them to semantically different classes (e.g., a cockatiel is reconstructed as a hamster), causing a pronounced shift in the logit-energy ratio that $\epsilon_{\text{logit}}$ detects.
These observations confirm that $\epsilon_{\text{feat}}$ and $\epsilon_{\text{logit}}$ capture complementary aspects of distributional shift, motivating their combination in the ensemble score of {\ours}.

\section{Complete results}
\label{app:sec:complete-results}

Tables \ref{tab:exp-imgnet-near-far-ood-app} present complete detection results on each OoD dataset under the near- and far-OoD setups in Sec.\ref{sec:exp-main} of the main text, corresponding to Table \ref{tab:exp-imgnet-near-far-ood}.
Tables \ref{tab:ablation-prob-feat-short-app} and \ref{tab:ablation-boosting-short-app} provide full detection results on each OoD dataset of the ablation studies in Sec.\ref{sec:exp-ablation} of the main text, corresponding to Tables \ref{tab:ablation-prob-feat-short} and \ref{tab:ablation-boosting-short}.

\begin{sidewaystable*}
    \centering
    \caption{Reconstruction examples of LMD \cite{liu2023unsupervised} from InD (ImageNet-1K) and OoD (OpenImage-O). A DM pretrained on ImageNet-1K with complex natural images can also successfully reconstruc those OoD images, leading to the weak performance of LMD.}
    \label{app:tab:ldm-results}
    \begin{tabular*}{\textheight}{ccc|ccc}
    \toprule
    \multicolumn{3}{c|}{InD (ImageNet-1K)} & \multicolumn{3}{c}{OoD (OpenImage-O)} \\
    Original & Masked & Inpainted & Original & Masked & Inpainted \\
    \midrule
    \begin{minipage}[b]{0.145\columnwidth}
		\centering
		\raisebox{-.5\height}{\includegraphics[width=\linewidth]{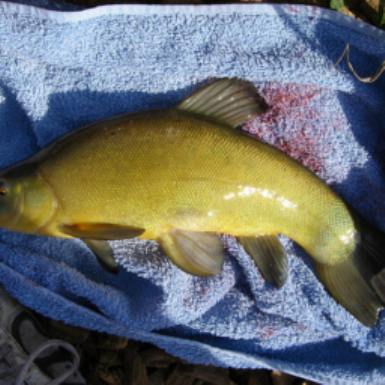}}
	\end{minipage} &
    \begin{minipage}[b]{0.145\columnwidth}
		\centering
		\raisebox{-.5\height}{\includegraphics[width=\linewidth]{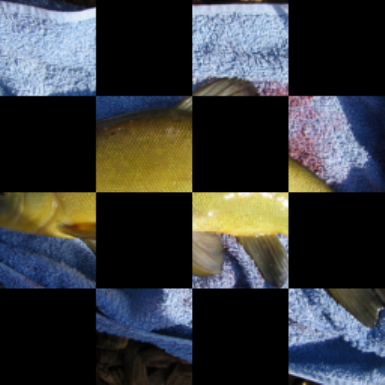}}
	\end{minipage} &
    \begin{minipage}[b]{0.145\columnwidth}
		\centering
		\raisebox{-.5\height}{\includegraphics[width=\linewidth]{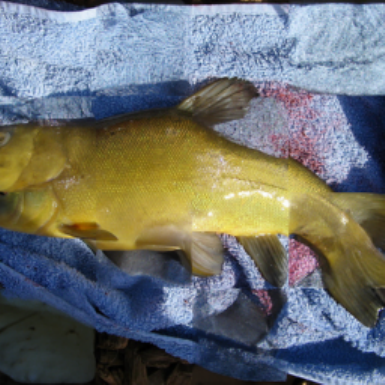}}
	\end{minipage} &
    \begin{minipage}[b]{0.145\columnwidth}
		\centering
		\raisebox{-.5\height}{\includegraphics[width=\linewidth]{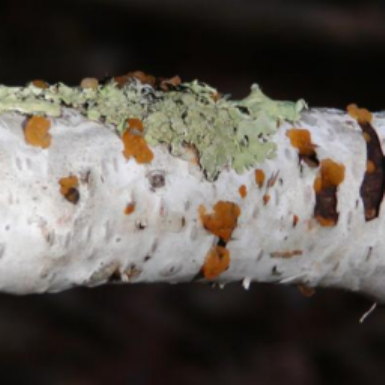}}
	\end{minipage} &
    \begin{minipage}[b]{0.145\columnwidth}
		\centering
		\raisebox{-.5\height}{\includegraphics[width=\linewidth]{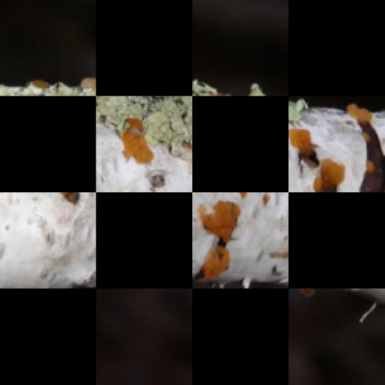}}
	\end{minipage} &
    \begin{minipage}[b]{0.145\columnwidth}
		\centering
		\raisebox{-.5\height}{\includegraphics[width=\linewidth]{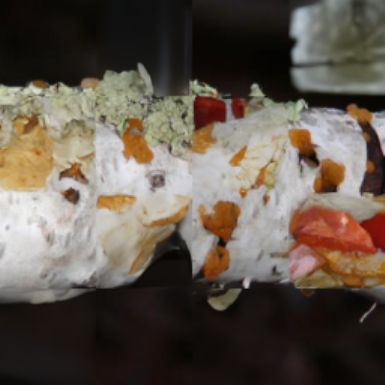}}
	\end{minipage} \\
    \midrule
    \begin{minipage}[b]{0.145\columnwidth}
		\centering
		\raisebox{-.5\height}{\includegraphics[width=\linewidth]{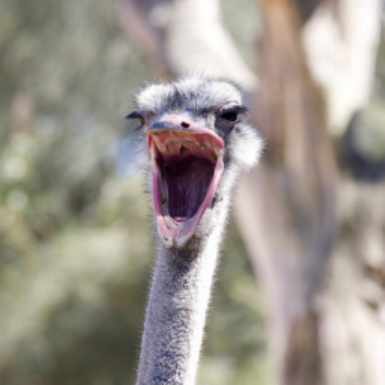}}
	\end{minipage} &
    \begin{minipage}[b]{0.145\columnwidth}
		\centering
		\raisebox{-.5\height}{\includegraphics[width=\linewidth]{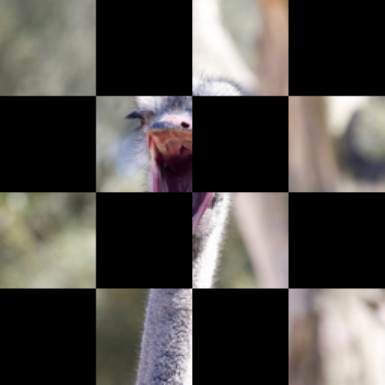}}
	\end{minipage} &
    \begin{minipage}[b]{0.145\columnwidth}
		\centering
		\raisebox{-.5\height}{\includegraphics[width=\linewidth]{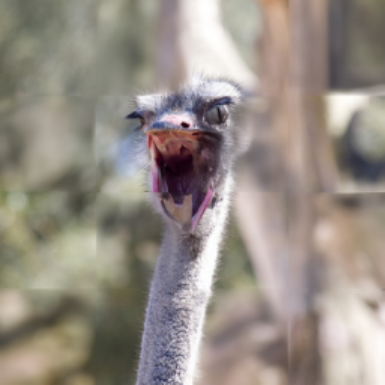}}
	\end{minipage} &
    \begin{minipage}[b]{0.145\columnwidth}
		\centering
		\raisebox{-.5\height}{\includegraphics[width=\linewidth]{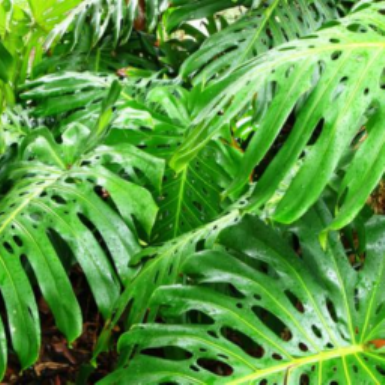}}
	\end{minipage} &
    \begin{minipage}[b]{0.145\columnwidth}
		\centering
		\raisebox{-.5\height}{\includegraphics[width=\linewidth]{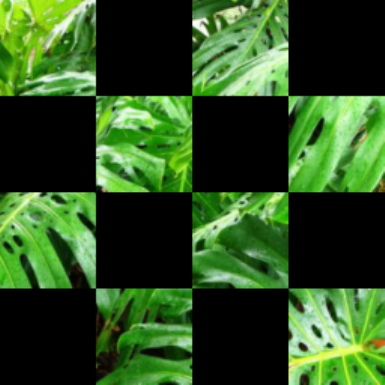}}
	\end{minipage} &
    \begin{minipage}[b]{0.145\columnwidth}
		\centering
		\raisebox{-.5\height}{\includegraphics[width=\linewidth]{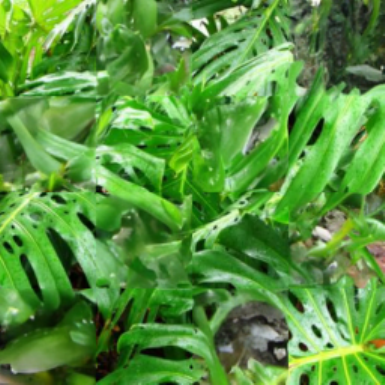}}
	\end{minipage} \\
    \midrule
    \begin{minipage}[b]{0.145\columnwidth}
		\centering
		\raisebox{-.5\height}{\includegraphics[width=\linewidth]{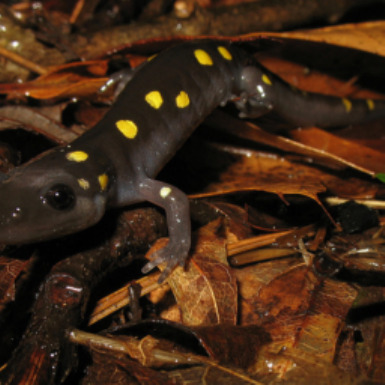}}
	\end{minipage} &
    \begin{minipage}[b]{0.145\columnwidth}
		\centering
		\raisebox{-.5\height}{\includegraphics[width=\linewidth]{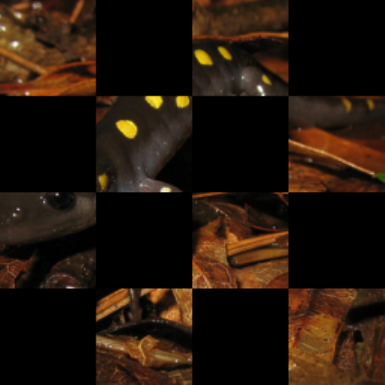}}
	\end{minipage} &
    \begin{minipage}[b]{0.145\columnwidth}
		\centering
		\raisebox{-.5\height}{\includegraphics[width=\linewidth]{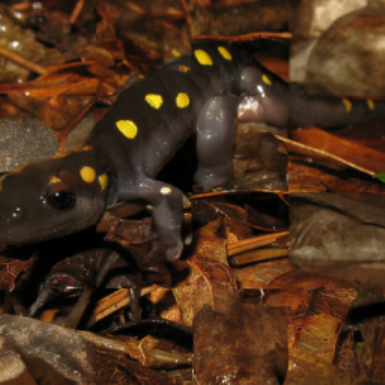}}
	\end{minipage} &
    \begin{minipage}[b]{0.145\columnwidth}
		\centering
		\raisebox{-.5\height}{\includegraphics[width=\linewidth]{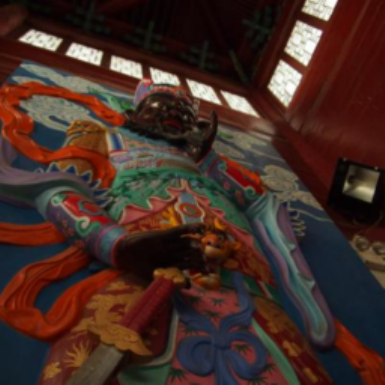}}
	\end{minipage} &
    \begin{minipage}[b]{0.145\columnwidth}
		\centering
		\raisebox{-.5\height}{\includegraphics[width=\linewidth]{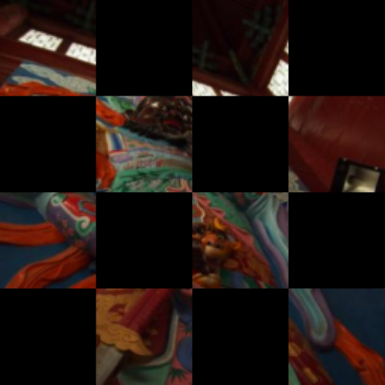}}
	\end{minipage} &
    \begin{minipage}[b]{0.145\columnwidth}
		\centering
		\raisebox{-.5\height}{\includegraphics[width=\linewidth]{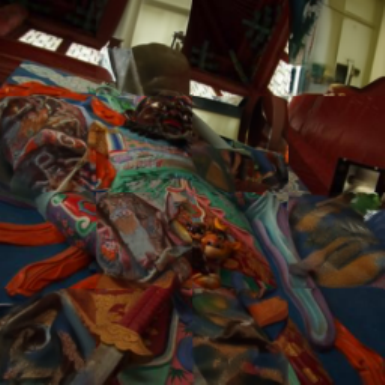}}
	\end{minipage} \\
    \bottomrule
    \end{tabular*}
\end{sidewaystable*}

\begin{sidewaystable*}
    \centering
    \caption{Visualization on the generated images w.r.t. different time steps $T$ in the reverse process. A small number of the sampling steps leads to low-quality images. Such examples also demonstrate that the underlying discrepancy between $\bm x$ and $\bm{\hat x}$ cannot be simply captured via perceptual metrics on their visual similarity.}
    \label{app:tab:ddr-step}
    \begin{tabular*}{\textheight}{ccccc|c}
    \toprule
    $T=20$ & $T=21$ & $T=22$ & $T=23$ & $T=24$ & original \\
    \midrule
    \begin{minipage}[b]{0.145\columnwidth}
		\centering
		\raisebox{-.5\height}{\includegraphics[width=\linewidth]{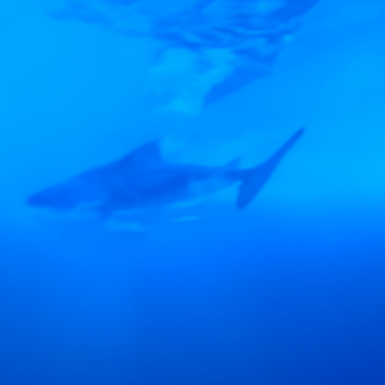}}
	\end{minipage} &
    \begin{minipage}[b]{0.145\columnwidth}
		\centering
		\raisebox{-.5\height}{\includegraphics[width=\linewidth]{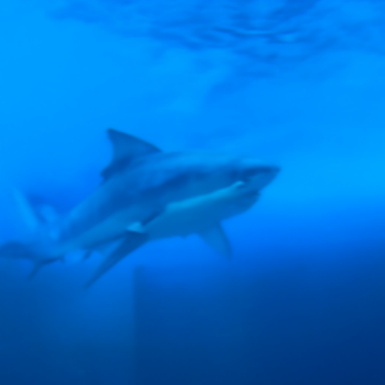}}
	\end{minipage} &
    \begin{minipage}[b]{0.145\columnwidth}
		\centering
		\raisebox{-.5\height}{\includegraphics[width=\linewidth]{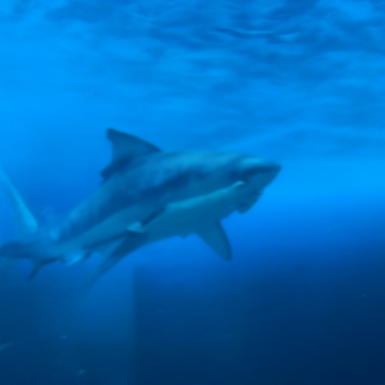}}
	\end{minipage} &
    \begin{minipage}[b]{0.145\columnwidth}
		\centering
		\raisebox{-.5\height}{\includegraphics[width=\linewidth]{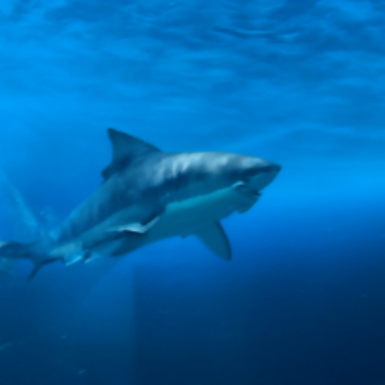}}
	\end{minipage} &
    \begin{minipage}[b]{0.145\columnwidth}
		\centering
		\raisebox{-.5\height}{\includegraphics[width=\linewidth]{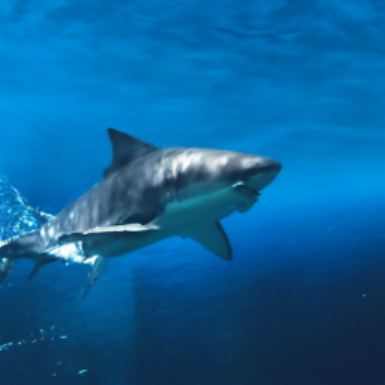}}
	\end{minipage} &
    \begin{minipage}[b]{0.145\columnwidth}
		\centering
		\raisebox{-.5\height}{\includegraphics[width=\linewidth]{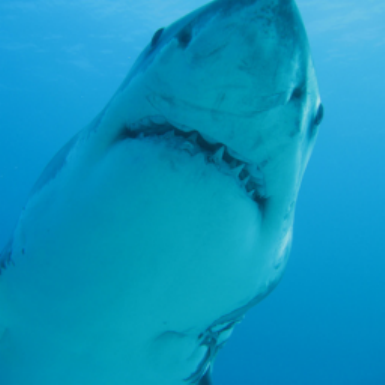}}
	\end{minipage} \\
    \midrule
    \begin{minipage}[b]{0.145\columnwidth}
		\centering
		\raisebox{-.5\height}{\includegraphics[width=\linewidth]{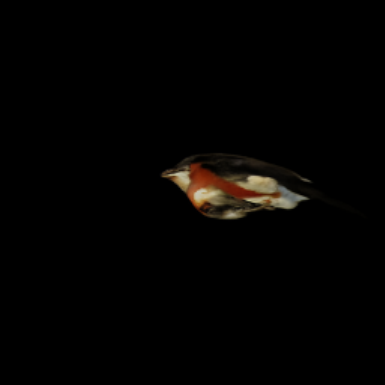}}
	\end{minipage} &
    \begin{minipage}[b]{0.145\columnwidth}
		\centering
		\raisebox{-.5\height}{\includegraphics[width=\linewidth]{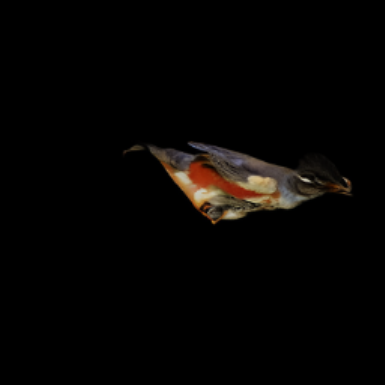}}
	\end{minipage} &
    \begin{minipage}[b]{0.145\columnwidth}
		\centering
		\raisebox{-.5\height}{\includegraphics[width=\linewidth]{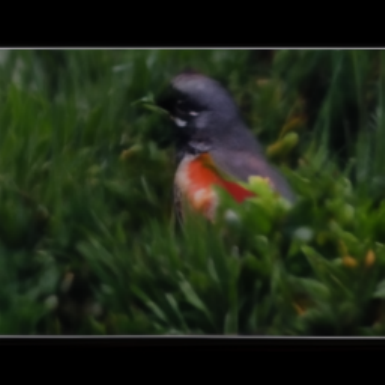}}
	\end{minipage} &
    \begin{minipage}[b]{0.145\columnwidth}
		\centering
		\raisebox{-.5\height}{\includegraphics[width=\linewidth]{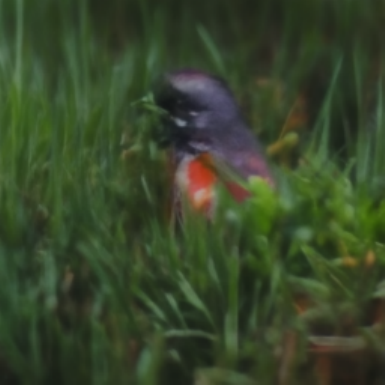}}
	\end{minipage} &
    \begin{minipage}[b]{0.145\columnwidth}
		\centering
		\raisebox{-.5\height}{\includegraphics[width=\linewidth]{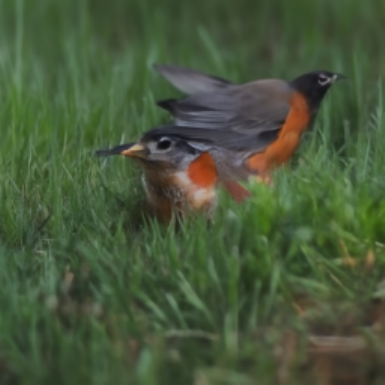}}
	\end{minipage} &
    \begin{minipage}[b]{0.145\columnwidth}
		\centering
		\raisebox{-.5\height}{\includegraphics[width=\linewidth]{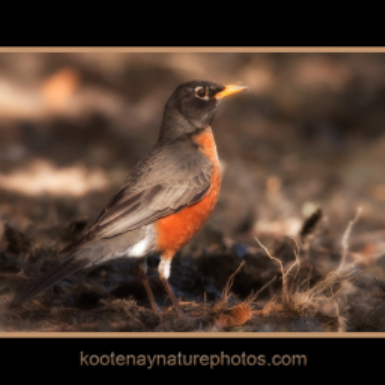}}
	\end{minipage} \\
    \midrule
    \begin{minipage}[b]{0.145\columnwidth}
		\centering
		\raisebox{-.5\height}{\includegraphics[width=\linewidth]{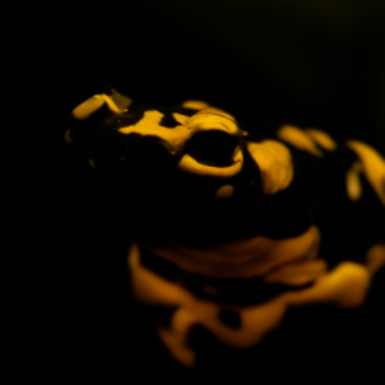}}
	\end{minipage} &
    \begin{minipage}[b]{0.145\columnwidth}
		\centering
		\raisebox{-.5\height}{\includegraphics[width=\linewidth]{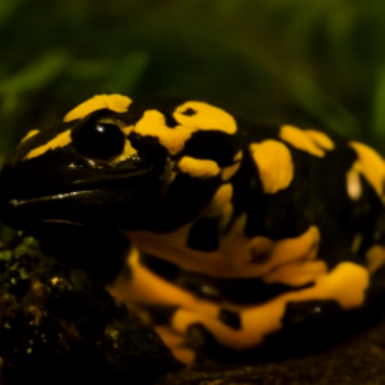}}
	\end{minipage} &
    \begin{minipage}[b]{0.145\columnwidth}
		\centering
		\raisebox{-.5\height}{\includegraphics[width=\linewidth]{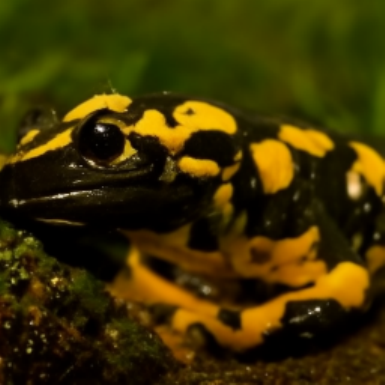}}
	\end{minipage} &
    \begin{minipage}[b]{0.145\columnwidth}
		\centering
		\raisebox{-.5\height}{\includegraphics[width=\linewidth]{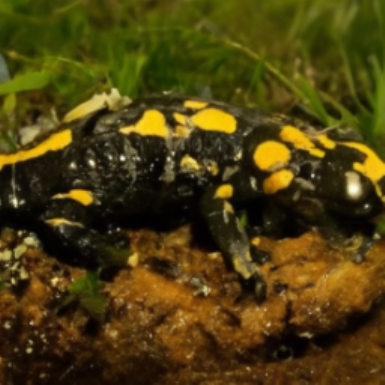}}
	\end{minipage} &
    \begin{minipage}[b]{0.145\columnwidth}
		\centering
		\raisebox{-.5\height}{\includegraphics[width=\linewidth]{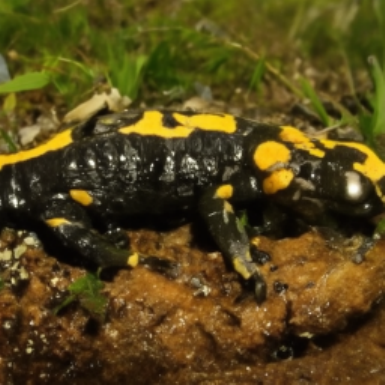}}
	\end{minipage} &
    \begin{minipage}[b]{0.145\columnwidth}
		\centering
		\raisebox{-.5\height}{\includegraphics[width=\linewidth]{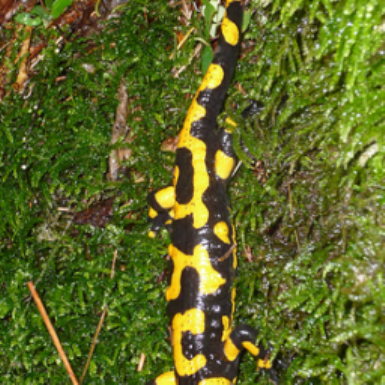}}
	\end{minipage} \\
    \bottomrule
    \end{tabular*}
\end{sidewaystable*}

\begin{sidewaystable*}
\centering
\caption{Visualization of complementary OoD samples from OpenImage-O.
\textbf{Left (Case~A)}: samples detected as OoD by the feature-level metric $\epsilon_{\rm feat}$ in {\ours} but \emph{missed} by the logic-level metric $\epsilon_{\rm logt}$, exemplifying covariate-shift.
\textbf{Right (Case~B)}: samples detected as OoD by the logic-level metric $\epsilon_{\rm logt}$ in {\ours} but \emph{missed} by the feature-level metric $\epsilon_{\rm feat}$, exemplifying concept-shift.
For each sample, the \emph{top row} shows the original input image and the \emph{bottom row} shows its DM-generated counterpart.}
    \label{app:tab:complementary-samples}
    \begin{tabular*}{\textheight}{lccc|ccc}
    \toprule
    & \multicolumn{3}{c|}{\textbf{Case A}: feature score detects OoD, logit score misses}
    & \multicolumn{3}{c}{\textbf{Case B}: logit score detects OoD, feature score misses} \\
    & Sample 1 & Sample 2 & Sample 3
    & Sample 1 & Sample 2 & Sample 3 \\
    \midrule
    Original &
    \begin{minipage}[b]{0.13\columnwidth}
        \centering
        \raisebox{-.5\height}{\includegraphics[width=\linewidth]{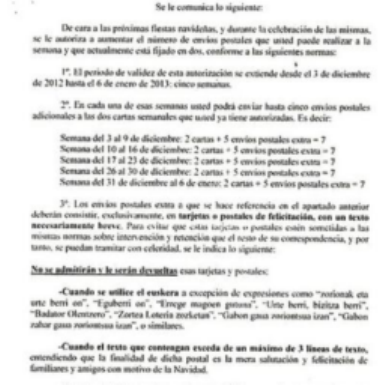}}
    \end{minipage} &
    \begin{minipage}[b]{0.13\columnwidth}
        \centering
        \raisebox{-.5\height}{\includegraphics[width=\linewidth]{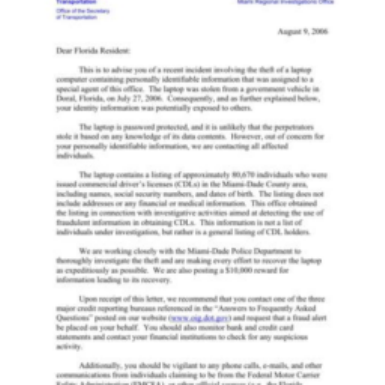}}
    \end{minipage} &
    \begin{minipage}[b]{0.13\columnwidth}
        \centering
        \raisebox{-.5\height}{\includegraphics[width=\linewidth]{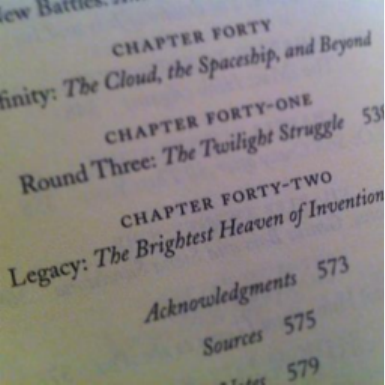}}
    \end{minipage} &
    \begin{minipage}[b]{0.13\columnwidth}
        \centering
        \raisebox{-.5\height}{\includegraphics[width=\linewidth]{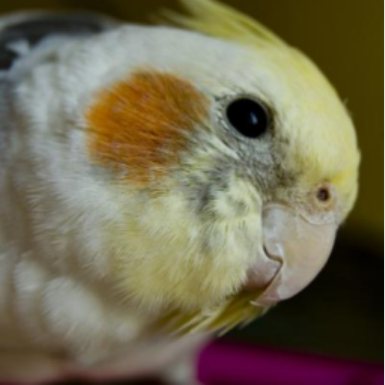}}
    \end{minipage} &
    \begin{minipage}[b]{0.13\columnwidth}
        \centering
        \raisebox{-.5\height}{\includegraphics[width=\linewidth]{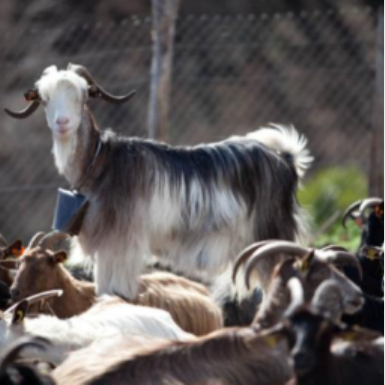}}
    \end{minipage} &
    \begin{minipage}[b]{0.13\columnwidth}
        \centering
        \raisebox{-.5\height}{\includegraphics[width=\linewidth]{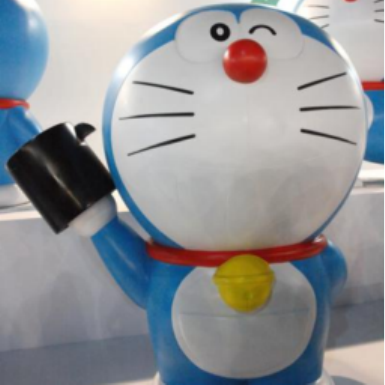}}
    \end{minipage} \\
    \midrule
    Reconstructed &
    \begin{minipage}[b]{0.13\columnwidth}
        \centering
        \raisebox{-.5\height}{\includegraphics[width=\linewidth]{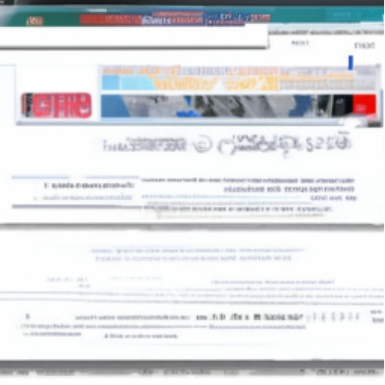}}
    \end{minipage} &
    \begin{minipage}[b]{0.13\columnwidth}
        \centering
        \raisebox{-.5\height}{\includegraphics[width=\linewidth]{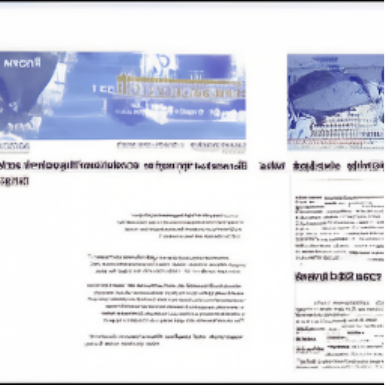}}
    \end{minipage} &
    \begin{minipage}[b]{0.13\columnwidth}
        \centering
        \raisebox{-.5\height}{\includegraphics[width=\linewidth]{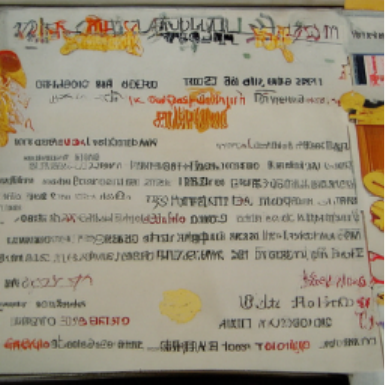}}
    \end{minipage} &
    \begin{minipage}[b]{0.13\columnwidth}
        \centering
        \raisebox{-.5\height}{\includegraphics[width=\linewidth]{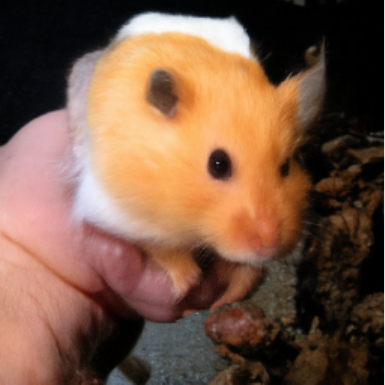}}
    \end{minipage} &
    \begin{minipage}[b]{0.13\columnwidth}
        \centering
        \raisebox{-.5\height}{\includegraphics[width=\linewidth]{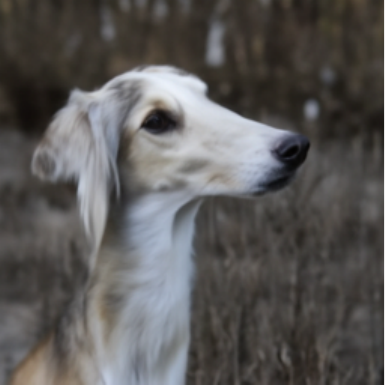}}
    \end{minipage} &
    \begin{minipage}[b]{0.13\columnwidth}
        \centering
        \raisebox{-.5\height}{\includegraphics[width=\linewidth]{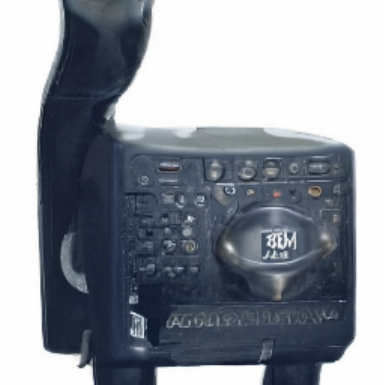}}
    \end{minipage} \\
    \bottomrule
    \end{tabular*}
\end{sidewaystable*}

\begin{table*}[t]
    \centering
    \caption{Results of a variety of OoD detection methods on {ImageNet-1K} with {ResNet50}. The {\bf best} and \underline{runner-up} results are highlighted with bond fonts and underlines, respectively.}
    \label{tab:exp-imgnet-near-far-ood-app}
    \resizebox{\textwidth}{!}{
    \begin{tabular}{c|cc cc cc|cc cc cc cc}
    \toprule
    \multirow{3}{*}{method} & \multicolumn{6}{c|}{Near-OoD} & \multicolumn{8}{c}{Far-OoD}\\
    & \multicolumn{2}{c}{NINCO} & \multicolumn{2}{c}{SSB-hard} & \multicolumn{2}{c|}{\bf Average} & \multicolumn{2}{c}{iNaturalist} & \multicolumn{2}{c}{Texture} & \multicolumn{2}{c}{OpenImage-O} & \multicolumn{2}{c}{\bf Average}\\
    \cmidrule{2-15}
    & FPR$\downarrow$ & AUROC$\uparrow$ 
    & FPR$\downarrow$ & AUROC$\uparrow$ 
    & FPR$\downarrow$ & AUROC$\uparrow$ 
    & FPR$\downarrow$ & AUROC$\uparrow$ 
    & FPR$\downarrow$ & AUROC$\uparrow$ 
    & FPR$\downarrow$ & AUROC$\uparrow$ 
    & FPR$\downarrow$ & AUROC$\uparrow$ \\
    \midrule
    \multicolumn{15}{c}{\it non-diffusion-based}\\
    MSP \cite{hendrycks2016baseline} & 76.39 & 79.94 & 85.01 & 72.08 & 80.70 & 76.01 &  52.82 & 88.39 & 66.28 & 80.43 & 64.05 & 84.85 & 61.05 & 84.56 \\
    Energy \cite{liu2020energy} & 77.61 & 79.69 & 84.42 & 72.08 & 81.02 & 75.89 & 53.76 & 90.62 & 52.46 & 86.72 & 57.69 & 89.02 & 54.64 & 88.79 \\
    GradNorm \cite{huang2021importance} & 73.89 & 74.06 & \underline{81.22} & 71.96 & 77.55 & 73.01 & 26.78 & 93.90 & 32.84 & 90.64 & 48.24 & 84.79 & 35.95 & 89.77 \\
    ReAct \cite{sun2021react} & 71.45 & 80.00 & 78.92 & 72.80 & 75.19 & 76.40 & 19.53 & 96.40 & 46.40 & 90.31 & 44.05 & 90.42 & 36.66 & 92.38 \\
    VRA \cite{xu2023vra} & 70.65 & \underline{81.53} & 84.81 & 70.05 & 77.73 & 75.79 & {\bf15.67} & {\bf97.14} & 21.67 & 95.64 & \underline{36.61} & {\bf92.94} & \underline{24.65} & \underline{95.24} \\
    KNN \cite{sun2022out} & 79.86 & 74.55 & 92.44 & 55.78 & 86.15 & 65.16 & 60.00 & 86.17 & 11.08 & 97.43 & 59.28 & 85.39 & 43.46 & 89.66 \\
    NNGuide \cite{park2023nearest} & \underline{69.21} & 81.45 & {\bf80.74} & \underline{73.18} & \underline{74.98} & \underline{77.32} & 25.61 & 95.15 & 27.73 & 92.31 & 40.27 & 91.82 & 31.20 & 93.10 \\
    MLS \cite{hendrycks2022scaling} & 76.64 & 80.40 & 84.39 & 72.50 & 80.52 & 76.45 & 50.81 & 91.15 & 54.96 & 86.39 & 57.93 & 89.13 & 54.57 & 88.89 \\
    KPCA \cite{fang2024kernel} & 77.49 & 76.34 & 89.68 & 61.99 & 83.58 & 69.16 & 49.84 & 89.40 & \underline{8.95} & \underline{98.16} & 54.74 & 87.10 & 37.85 & 91.55 \\
    Mahala++ \cite{mullermahalanobis++} & 75.57 & 79.23 & 89.90 & 64.95 & 82.74 & 72.09 & 50.17 & 90.62 & {\bf5.73} & {\bf98.78} & 46.70 & 90.40 & 34.20 & 93.27 \\
    \midrule
    \multicolumn{15}{c}{\it diffusion-based}\\
    DDPM-OOD \cite{graham2023denoising} & 92.65 & 49.57 & 94.43 & 51.78 & 93.54 & 50.68 & 94.70 & 50.10 & 94.61 & 38.57 & 94.11 & 47.45 & 94.47 & 45.38 \\
    LMD \cite{liu2023unsupervised} & 94.34 & 50.27 & 94.51 & 53.50 & 94.42 & 51.88 & 94.65 & 60.04 & 80.23 & 60.36 & 93.50 & 51.36 & 89.46 & 57.25 \\
    DiffPath \cite{heng2024out} & 88.65 & 59.35 & 95.70 & 45.90 & 92.18 & 52.63 & 90.50 & 59.42 & 78.14 & 71.57 & 90.70 & 57.98 & 86.45 & 62.99 \\
    DiffPathV2 \cite{Abdi_2025_ICCV} & 90.69 & 52.75 & 92.42 & 52.65 & 91.56 & 52.70 & 95.57 & 45.30 & 88.69 & 58.16 & 95.35 & 49.81 & 93.20 & 51.09 \\
    DiffGuard \cite{gao2023diffguard} & 81.29 & 75.44 & 85.97 & 65.87 & 83.63 & 70.65 & 71.23 & 85.81 & 55.51 & 85.41 & 74.80 & 82.32 & 67.18 & 84.51 \\
    \cmidrule{2-15}
    \rowcolor{tabgray} {\ours} & {\bf66.75} & {\bf83.04} & 81.31 & {\bf74.16} & {\bf74.03} & {\bf78.60} & \underline{17.96} & \underline{96.69} & 9.63 & 97.97 & {\bf35.02} & \underline{92.86} & {\bf20.87} & {\bf95.84} \\
    \bottomrule
    \end{tabular}}
\end{table*}

\begin{table*}[t]
    \centering
    \caption{The individual effects of different alternatives of the metrics in the feature space and logit space of {\ours}. Detection results on each OoD dataset are reported.}
    \label{tab:ablation-prob-feat-short-app}
    \resizebox{\textwidth}{!}{
    \begin{tabular}{c|cc cc cc cc|cc}
    \toprule
    \multirow{3}{*}{Score} & \multicolumn{8}{c|}{OoD data sets} & \multicolumn{2}{c}{\multirow{2}{*}{\bf AVERAGE}}\\
    & \multicolumn{2}{c}{Species} & \multicolumn{2}{c}{iNaturalist} & \multicolumn{2}{c}{OpenImage-O} & \multicolumn{2}{c|}{ImageNet-O} \\
    \cmidrule{2-11}
    & FPR$\downarrow$ & AUROC$\uparrow$ & FPR$\downarrow$ & AUROC$\uparrow$ &
    FPR$\downarrow$ & AUROC$\uparrow$ & FPR$\downarrow$ & AUROC$\uparrow$ & FPR$\downarrow$ & AUROC$\uparrow$ \\
    \midrule
    $\epsilon_{\rm feat}$ & 82.44 & 72.42 & 42.12 & 90.78 & 58.29 & 81.69 & 73.00 & 78.28 & 63.96 & 80.79 \\
    $\epsilon_{\rm logt}$ & 82.83 & 71.38 & 62.42 & 88.42 & 63.29 & 87.29 & 99.95 & 40.12 & 77.12 & 71.80 \\
    Ensemble & {\bf67.82} & {\bf80.17} & {\bf17.96} & {\bf96.69} & {\bf35.02} & {\bf92.86} & {86.80} & {69.08} & {\bf51.90} & {\bf84.70} \\
    \midrule
    $\epsilon_{\rm feat}$ ($\ell_2$) & 83.02 & 72.02 & 47.48 & 89.10 & 65.73 & 82.18 & 81.15 & 70.61 & 69.35 & 78.48 \\
    $\epsilon_{\rm feat}$ (cosine) & 83.02 & 72.02 & 47.48 & 89.10 & 65.73 & 82.18 & 81.15 & 70.61 & 69.35 & 78.48 \\
    \bottomrule
    \end{tabular}}
\end{table*}

\begin{table*}[t]
    \centering
    \caption{Ablation studies on the subspace-based representation refinement strategy of {\ours} with results on each OoD dataset.}
    \label{tab:ablation-boosting-short-app}
    \resizebox{\textwidth}{!}{
    \begin{tabular}{c|cc cc cc cc|cc}
    \toprule
    \multirow{3}{*}{Implementation} & \multicolumn{8}{c|}{OoD data sets} & \multicolumn{2}{c}{\multirow{2}{*}{\bf AVERAGE}}\\
    & \multicolumn{2}{c}{Species} & \multicolumn{2}{c}{iNaturalist} & \multicolumn{2}{c}{OpenImage-O} & \multicolumn{2}{c|}{ImageNet-O} \\
    \cmidrule{2-11}
    & FPR$\downarrow$ & AUROC$\uparrow$ & FPR$\downarrow$ & AUROC$\uparrow$ &
    FPR$\downarrow$ & AUROC$\uparrow$ & FPR$\downarrow$ & AUROC$\uparrow$ & FPR$\downarrow$ & AUROC$\uparrow$ \\
    \midrule
    w/o refine & 71.05 & 77.16 & 36.47 & 92.96 & 48.58 & 89.63 & 92.25 & 62.28 & 62.09 & 80.51 \\
    Refine on ${\bm h}_{\bm{x}},{\bm z}_{\bm{x}},{\bm h}_{\bm{\hat x}},{\bm z}_{\bm{\hat x}}$ & 72.60 & 75.26 & 40.82 & 91.86 & 56.72 & 85.77 & 89.20 & 66.57 & 64.83 & 79.87 \\
    Refine on ${\bm h}_{\bm{x}},{\bm z}_{\bm{x}}$ & 82.58 & 71.42 & 74.89 & 77.46 & 83.31 & 66.30 & 89.55 & 54.56 & 82.58 & 67.43 \\
    Refine on ${\bm h}_{\bm{\hat x}},{\bm z}_{\bm{\hat x}}$ & {\bf67.82} & {\bf80.17} & {\bf17.96} & {\bf96.69} & {\bf35.02} & {\bf92.86} & {\bf86.80} & {\bf69.08} & {\bf51.90} & {\bf84.70} \\
    \bottomrule
    \end{tabular}}
\end{table*}

\end{appendices}

\end{document}